\documentclass[]{gtech}
\usepackage{amssymb}
\usepackage{multirow}
\usepackage{bigdelim}
\usepackage{todonotes}
\usepackage{longtable}
\usepackage{tabularray}
\usepackage{wrapfig}
\usepackage[most]{tcolorbox} 
\usepackage{xcolor}
\usepackage{url}
\usepackage{algorithm}  
\usepackage{algpseudocode} 
\usepackage[toc,page]{appendix}
\usepackage{multirow}
\usepackage[normalem]{ulem}
\usepackage{adjustbox}
\usepackage{booktabs}  
\usepackage{natbib}
\usepackage{colortbl}  
\usepackage{subcaption}
\usepackage{pifont}
\usepackage{amsmath}
\usepackage{amsfonts}       
\usepackage{multirow}
\usepackage{mathrsfs}
\usepackage{tabularx}
\usepackage[utf8]{inputenc} 
\usepackage[T1]{fontenc}    
\usepackage{hyperref}       
\usepackage{url}            
\usepackage{booktabs}       
\usepackage{nicefrac}       
\usepackage{microtype}      
\usepackage{xcolor}         
\definecolor{ourreward}{HTML}{32bbee}
\definecolor{sparsereward}{HTML}{ed9a81}
\usepackage{graphicx}
\usepackage{wrapfig}
\usepackage{cleveref}

\usepackage{bbding}

\usepackage{multicol}

\usepackage{CJK}

\usepackage{bbm}
\useunder{\uline}{\ul}{}

\title{%
  \texorpdfstring{%
    \raisebox{-0.23\height}{\includegraphics[scale=0.03]{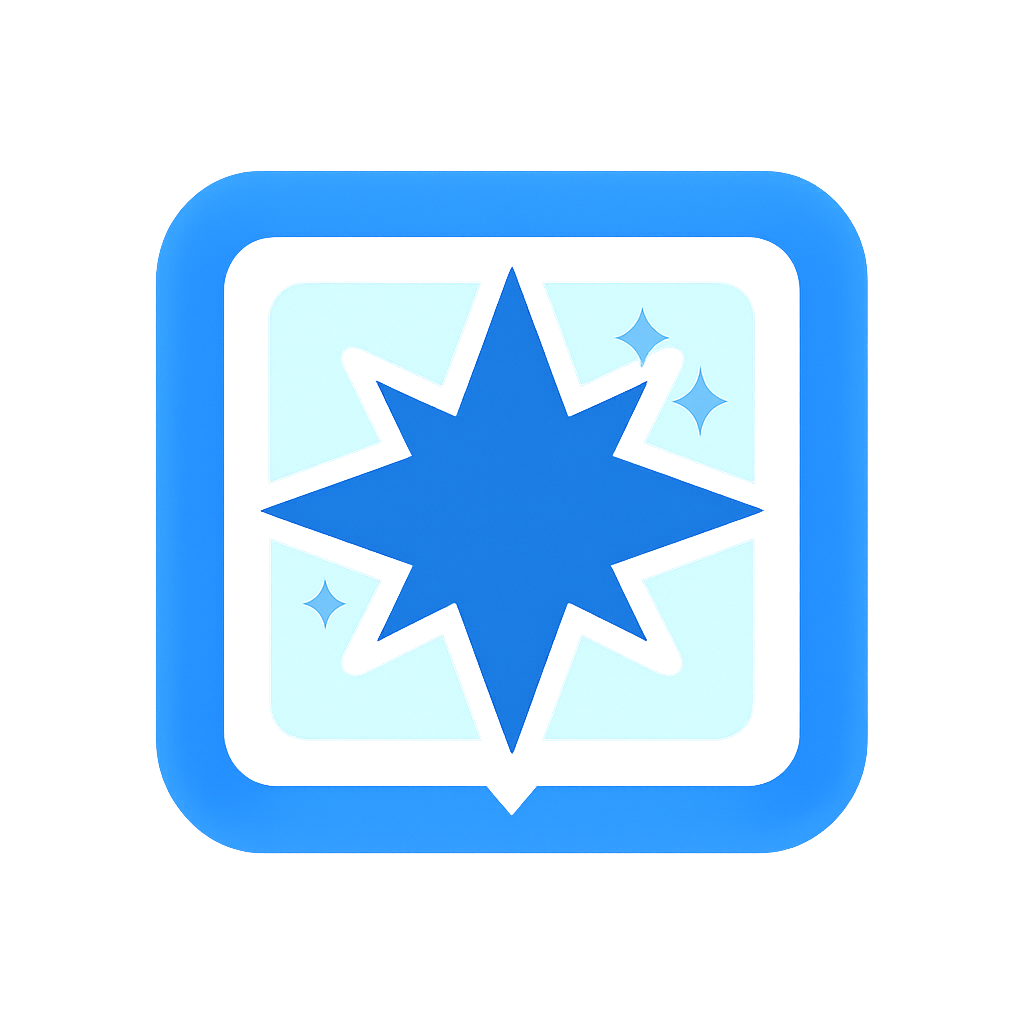}}%
    ~UI-Venus-2 Technical Report%
  }{UI-Venus-2 Technical Report}%
}

\author{Venus Team, Ant Group}

\abstract{
Multimodal GUI agents have emerged as a promising paradigm for digital task automation, yet transitioning from benchmark-oriented models to dependable real-world applications remains challenging due to limited environment coverage, brittle task construction, and unreliable reward verification. In this work, we present \textbf{UI-Venus-2}, a general-purpose foundation GUI agent designed to operate across mobile, web, and desktop environments through a unified closed-loop reasoning–action framework. 
To bridge the gap toward practical deployment, we jointly scale three critical dimensions: (1) Environments, expanding coverage to 170+ multilingual mobile apps and native desktop OS; (2) Tasks, employing a deep-research pipeline for function-grounded instruction generation; and (3) Verification, adopting trace- and sample-level evaluators with visual keypoints and multi-model voting to ensure reliable RL signals for training.
Furthermore, we integrate safety-aware mechanisms to ensure controlled execution of consequential actions. By offering a capable, efficient, and open-source foundation, UI-Venus-2 advances the field toward more generalizable, verifiable, and self-reflective agents for real-world applications.
}

\gtechdata[Code]{\url{https://github.com/inclusionAI/UI-Venus}}
\gtechdata[Model]{\url{https://huggingface.co/collections/inclusionAI/ui-venus}}
\gtechdata[Project]{\url{https://ui-venus.github.io/UI-Venus-2}}

\begin{document}
\maketitle

\renewcommand{\thefootnote}{}


\begin{figure}[htbp]
	\centering
	\includegraphics[width=\textwidth]{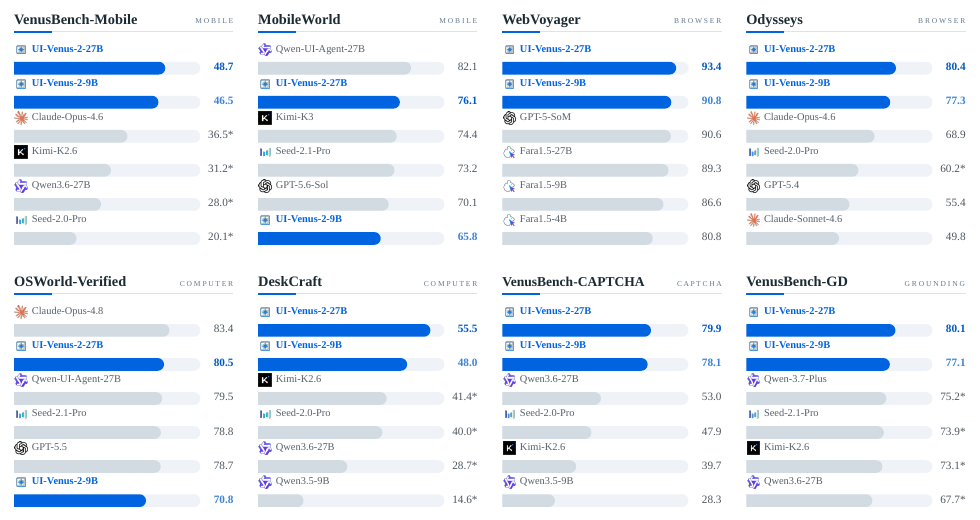}
	\caption{\textbf{Performances of UI-Venus-2 on GUI-agent benchmarks.} Each panel compares UI-Venus-2-27B and UI-Venus-2-9B with some selected strong baselines.
    We favor standalone end-to-end systems evaluated on the closest available task subset and step budget; source-reported action scaffolds may still differ. 
    MobileWorld uses GUI-only success rate on 117 tasks with 50 steps, WebVoyager uses the refreshed 595-task split, Odysseys uses average rubric score over 200 tasks, VenusBench-CAPTCHA uses micro Pass@1 over all 219 examples, and VenusBench-GD uses English-instruction micro-average accuracy. ``*'' denotes the results are reproduced by us.}
	\label{fig:performance_uivenus}
\end{figure}

\section{Introduction}

The ability to autonomously operate digital devices from natural-language instructions has long been a central goal of artificial intelligence. With the rapid development of Multimodal Large Language Models (MLLMs)~\cite{anthropic2024cuda,bai2025qwen25vltechnicalreport,zhu2025internvl3exploringadvancedtraining,glm-4.5v,seed1.8,Qwen3-VL,team2025every1,ai2025ming,team2025every2}, GUI agents~\cite{gu2025ui,guig2,ye2025mobile,yan2025step,zhou2025mai,wang2025opencuaopenfoundationscomputeruse,liu2024autoglm,hai2025holo2modelfamily,cheng2026openmobile,zhang2026omegausebuildinggeneralpurposegui} have emerged as a promising interface between high-level user intent and concrete digital execution. Rather than depending on platform-specific APIs or handcrafted workflows, these agents perceive rendered interfaces and interact through human-like actions such as clicking, typing, scrolling, and keyboard operations. This visual, interface-first paradigm makes it possible to automate applications for which structured APIs are unavailable or incomplete, and has consequently attracted substantial attention from both academia and industry.

Despite rapid progress in GUI grounding and navigation, moving from benchmark-oriented agents to dependable real-world applications requires more than improving the accuracy of the next predicted action. Recent user-centric evaluations, such as VenusBench-Mobile~\cite{venusbenchmobile2} and OSWorld~\cite{xie2024osworld}, further expose this gap: app-centric and task-homogeneous benchmarks can obscure capability failures, while agents remain brittle under realistic environment variations. 
First, environment coverage must extend beyond a small collection of applications to multilingual mobile ecosystems, dynamic websites, and full desktop operating systems. Second, scaling interactive environments also requires scalable task construction: generated queries must be grounded in the actual functionality of an application and remain executable under changing interface states. 
Finally, reinforcement learning is only as reliable as the data quality verifier that supplies its reward. A coarse or brittle verifier may confuse partial progress with true completion, overlook task-critical visual evidence, or expose reward signals that can be exploited by the policy. 
These challenges make environment coverage, task quality, and data verification tightly coupled components of a practical GUI-agent system. UI-Venus~\cite{gu2025ui} and UI-Venus-1.5~\cite{team2026ui} established a unified end-to-end framework for GUI grounding and navigation, progressively improving the agent through large-scale GUI mid-training, offline and online reinforcement learning, and the integration of specialized capabilities into a single model. Building on this foundation, we take a further step toward general-purpose computer use by systematically scaling the environments, tasks, and feedback available to the agent. 

In this work, We present \textbf{UI-Venus-2}, a general-purpose foundation GUI agent designed to operate across mobile applications, web platforms, and desktop operating systems. Rather than treating broader interaction coverage as an isolated scaling problem, UI-Venus-2 follows a unified closed-loop reasoning--action paradigm, in which the agent observes the current interface, reasons about the task state, executes an action, and incorporates subsequent environmental feedback into its next decision.

To develop {UI-Venus-2}, our key design principle is to scale environment, task distribution and data verification jointly. 
We substantially expand the executable environment pool to cover diverse mobile, web, and desktop scenarios, while developing task construction pipelines that produce executable and function-grounded tasks at scale. To support reliable reinforcement learning, we further introduce stronger both trace-level and step-level verification that evaluates whether the intended goal has actually been achieved rather than merely detecting superficial progress. 
Beyond broad environment and task coverage, end-to-end autonomy also requires handling CAPTCHAs in login, registration, and other workflows. This last-mile capability further supports data scaling by preventing verification gates from stalling trajectory collection and blocking downstream states; targeted CAPTCHA data, in turn, strengthens the capability.
In parallel, we incorporate safety-aware mechanisms to ensure that increased operational capability is accompanied by more reliable control over potentially consequential actions.

Moreover, we evaluate UI-Venus-2 across representative mobile, web, and computer-use settings. Figure~\ref{fig:performance_uivenus} provides an overview of the evaluation, covering tasks such as navigation, grounding, and CAPTCHA solving, while detailed experimental results and ablations are presented in the following sections. Collectively, UI-Venus-2 advances the UI-Venus family from an action-centric navigation agent toward a broader, verifiable, and self-reflective foundation agent for real-world computer use. Our contributions are summarized as follows:
\begin{itemize}
\item \textbf{Scaled multilingual mobile-use environments with various and reliable task generation.} We substantially expand the executable mobile environment to cover a broader range of both 100+ Chinese and 70+ English language applications. To generate high-quality tasks at this scale, we introduce a query-generation strategy based on deep-research that grounds task queries in the available application functionality, improving the accuracy, validity, and executability of generated instructions.
\item \textbf{Computer-use capability built from the ground up.} We extend the UI-Venus family beyond its previous mobile- and web-centered scope by constructing dedicated desktop operating-system capabilities from scratch. Through computer-use data collection and task-specific training, UI-Venus-2 learns to perceive, reason, and act in desktop environments, thereby supporting mobile, web, and OS interaction within a unified end-to-end agent.
\item \textbf{Trace-level and Sample-level verification.} We introduce a more precise verification mechanism that evaluates task completion using task-relevant visual keypoints rather than relying solely on a coarse holistic judgment of the final interface. We further aggregate judgments from multiple heterogeneous models through voting, improving verification robustness, reducing single-judge bias, and making the reward signal less susceptible to reward hacking.
\item \textbf{Open-sourced GUI Agent.} We publicly release the full-parameter weight,  and evaluation infrastructure of UI-Venus-2, which achieves almost state-of-the-art performance among models of comparable scale across multiple GUI benchmarks. By providing a highly capable yet computationally efficient foundation, we aim to lower the barrier to entry for GUI agent research, facilitate reproducible studies in reinforcement learning and verification, and accelerate community-driven innovation toward more generalizable and reliable computer-use agents.
\end{itemize}

\section{Methodology}

\begin{figure}[htbp]
	\centering
	\includegraphics[width=0.9\textwidth]{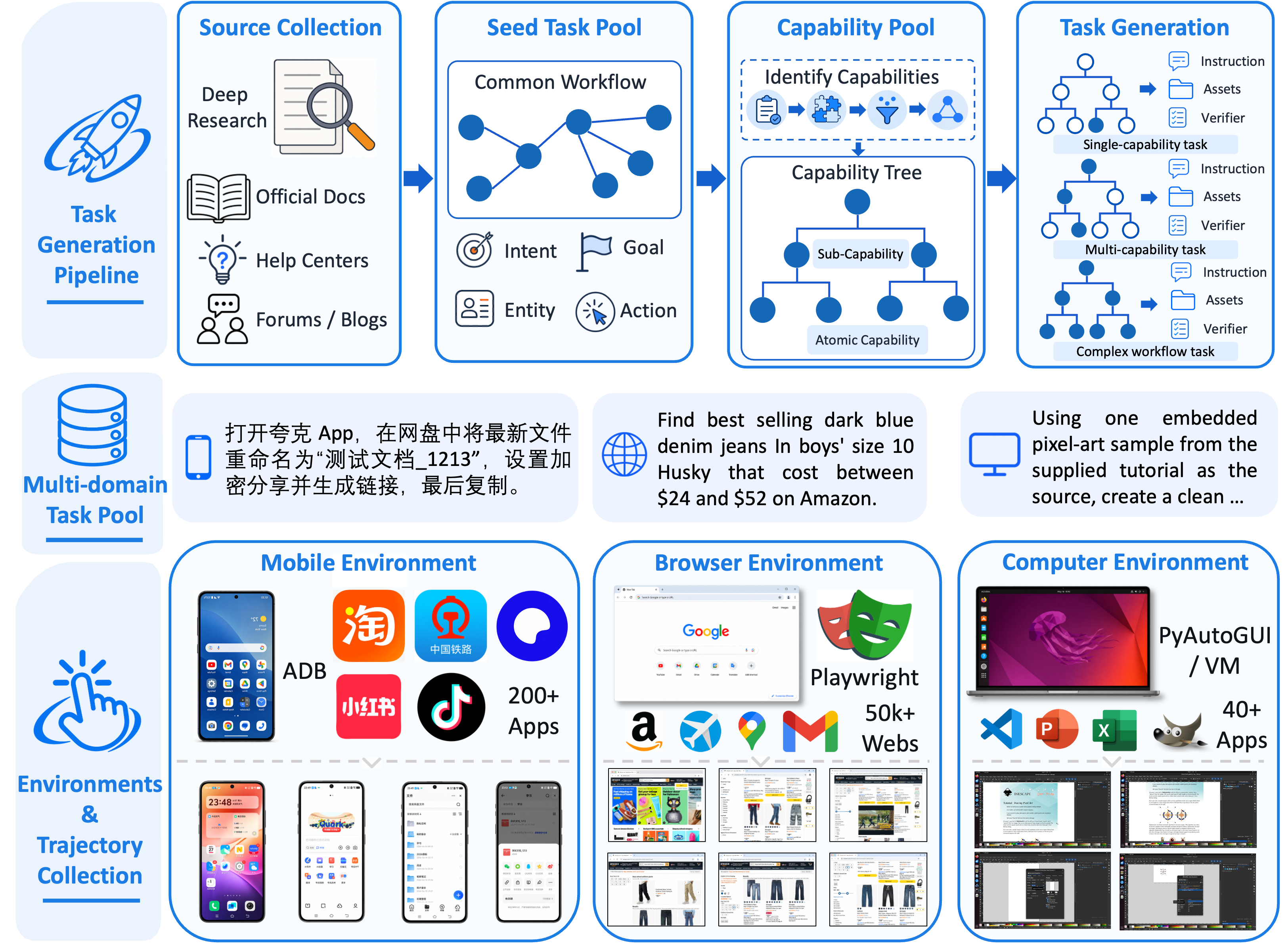}
	\caption{\textbf{System Overview of UI-Venus-2}. The figure illustrates the task generation and trajectory collection process of UI-Venus-2. Diverse tasks are constructed to form a multi-domain task pool, and interaction trajectories are collected across mobile, browser, and computer environments, covering a broad range of real-world applications, websites, and desktop software.}
	\label{fig:venus20_overview}
\end{figure}

\subsection{System overview}
UI-Venus-2 is a general-purpose foundation GUI agent that combines visual perception and advanced reasoning with reinforcement learning-driven interaction capabilities, enabling autonomous operation across diverse digital environments including mobile applications, desktop operating systems, and web platforms.
Given a natural language instruction, the model operates through a unified reasoning–action framework: it observes the rendered interface images, interprets the current visual context, translates high-level user intent into executable GUI actions, and continuously adapts its decisions based on environmental feedback until the task is completed. 

To support this interaction paradigm, UI-Venus-2 is trained through a unified pipeline that combines mid-training, offline reinforcement learning, and OPD, with the mid-training and RL components following UI-Venus-1.5~\cite{team2026ui}. This training process equips the model with strong GUI understanding, long-horizon planning, precise action grounding, and adaptive error recovery, enabling it to handle complex and dynamic workflows. We leave detailed descriptions of each training stage to the following sections.

\textbf{Action Spaces:}
When designing the action space for UI-Venus-2, we additionally design some actions of desktop to better interact with os-use instructions, as shown in the appendix.

\textbf{Model Init:}
We initialize our training pipeline from strong open-source multimodal foundation models, Qwen3.5-9B~\cite{qwen3.5} and Qwen3.6-27B~\cite{qwen3.6-27b}, which provide strong priors in visual understanding, multimodal reasoning, and instruction following. Rather than optimizing for CAPTCHA solving as an isolated capability, our framework aims to develop a broader spectrum of visual navigation and computer-use capabilities through a unified mixture of {Grounding, CAPTCHA, Mobile, Web, and Computer tasks}. These task families are deliberately complementary: Grounding provides fine-grained spatial perception, CAPTCHA offers controlled and verifiable interaction supervision, while Mobile, Web, and Computer tasks constitute the primary sources of navigation-oriented experience in realistic interactive environments. This unified formulation allows knowledge acquired from different domains to be mutually reinforcing rather than learned as independent task-specific skills.

\begin{figure}[htbp]
	\centering
	\includegraphics[width=0.9\textwidth]{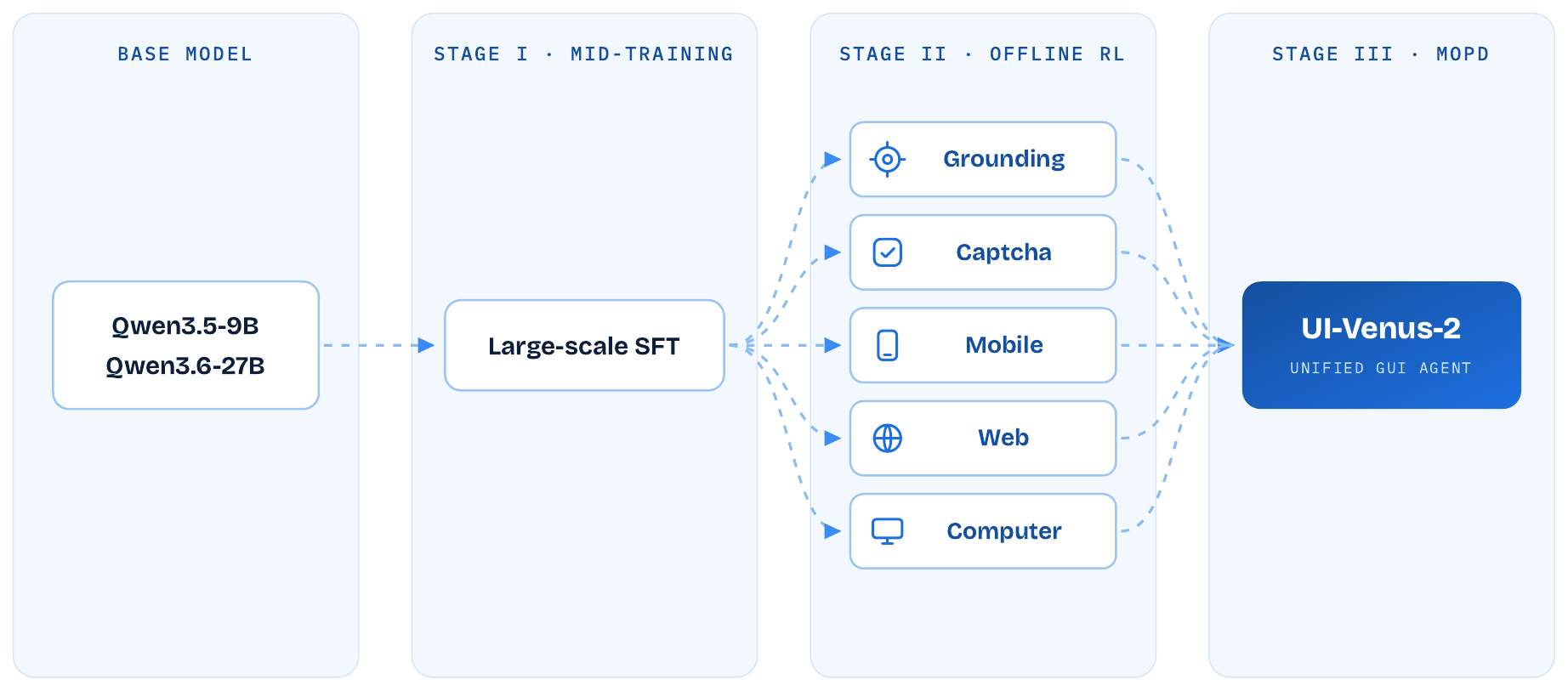}
	\caption{\textbf{The Three-Stage Pipeline of UI-Venus-2}.  Following the overall training recipe of UI-Venus-1.5, UI-Venus-2 starts with large-scale trajectory-based mid-training to inject GUI interaction knowledge. The resulting model is then optimized independently for each domain using step-level Offline-RL, covering Grounding, CAPTCHA, Mobile, Web, and Computer tasks. Finally, the domain-specialized models are consolidated into the final UI-Venus-2 model via multi-teacher on-policy distillation.}
	\label{fig:venus15_pipeline}
\end{figure}

\subsection{Stage I: Multimodal Mid-Training}
We first perform multimodal mid-training on a large-scale and heterogeneous mixture of synthetic and interaction-oriented data, with {Mobile, Web, and OS navigation tasks constituting the dominant component of the training corpus}. To obtain diverse and controllable navigation experience, we simulate a broad range of interactive environments spanning different application interfaces, webpages, and operating-system contexts, and collect executable trajectories across these environments. Task queries are constructed through a complementary combination of queries conditioned on curated query seeds, thereby balancing realism, linguistic diversity, and task coverage. The resulting trajectories are further validated through a {human--discriminator collaborative verification process}, which combines human inspection with automated trajectory-level assessment to filter invalid, ambiguous, or low-quality interactions. 
Detailed procedures for environment simulation, query construction, trajectory collection, and quality control are provided in the \textit{Data Preparation} section.

\subsection{Stage II: Offline Reinforcement Learning}
Starting from the mid-trained models, we further perform offline reinforcement learning with task-specific supervision tailored to different interaction scenarios. For {Mobile, OS, and Web environments}, we construct large-scale {step-level RL trajectories}, providing supervision at individual interaction steps to optimize state-aware action selection, multi-step navigation, transition consistency, and execution reliability. For {CAPTCHA and Grounding}, instead we use our programmatic synthesis framework to embed verified CAPTCHA instances and grounding targets into realistic {web-page and app background interfaces}, allowing the model to learn robust target identification and precise spatial localization under realistic visual clutter. 
Moreover, the generation process provides reliable action-level correctness and controllable difficulty, enabling dense and scalable RL supervision that would be difficult to obtain from naturally collected interaction data alone. This joint optimization consequently strengthens navigation robustness, spatial accuracy, and action execution consistency across domains.


\subsection{Stage III: Multi-teacher On-policy Distillation}
\label{sec:model-merge}

Finally, we employ {Multi-teacher On-policy Distillation (MOPD)}~\cite{xiao2026mimo,yan2026maga} to consolidate capabilities acquired from heterogeneous task distributions and training stages. This stage aims to preserve the broad multimodal reasoning capabilities inherited from the foundation models while integrating specialized skills from {Grounding and CAPTCHA} with the navigation expertise developed from {Mobile, Web, and Computer tasks}. The resulting model is not merely a collection of task-specific competencies, but a unified policy where spatial grounding supports element selection, verified interaction strengthens fine-grained execution, and navigation experience enables long-horizon composition.

To mitigate the interference that arises from directly merging independently trained experts, we adopt an on-policy distillation paradigm where teachers score trajectories sampled by the student itself. However, vanilla OPD applies uniform token-level supervision across the entire response, treating reasoning traces and executable actions alike. This is poorly suited to GUI agents because the action is the sole interface through which the agent interacts with and changes the environment. Although action tokens directly determine execution, they oconly a small portion of the response and may therefore receive insufficient supervision. Moreover, each action has an explicit structure consisting of an action type and its type-specific parameters. We therefore augment OPD with structured action-aware supervision and teacher-side hints, concentrating the distillation signal on the executable behaviors that govern environment interaction.

\paragraph{Structured Action-Aware Distillation.}
A GUI response typically contains a reasoning trace followed by a structured action, but only the action is exposed to the environment and determines the resulting state transition. An error in this short action span can invalidate an otherwise coherent response and move the agent away from the intended trajectory. Furthermore, the action exhibits internal dependency: the action type selects the parameter schema, and parameters are meaningful only under the corresponding type. Treating all response tokens uniformly overlooks this asymmetry between reasoning and execution.

We therefore condition the token-level distillation signal on the correctness of the student action while accounting for this structural dependency. Specifically: (1) if the complete action is correct, we suppress its distillation signal as no correction is needed; (2) if the action type is correct but the parameters are not, we strengthen supervision over the action span; (3) if the action type is incorrect, we emphasize the type tokens and mask the downstream parameters, since their meaning depends on the predicted type. For example, an incorrect click location can still be corrected when the action type is \texttt{Click}, whereas the parameters of an incorrectly predicted \texttt{Scroll} action are semantically irrelevant. This adaptive weighting directs supervision toward the executable component requiring correction rather than allocating it solely according to token frequency.

\paragraph{Teacher-Side Action-Type Conditioning.}
During training, the routed teacher scores the response sampled by the student. To provide more precise supervision on structured actions, we append a hint $h(z^*)$ regarding the correct action type to the teacher prompt. Crucially, this hint is never included in the student prompt and is unavailable at inference time; it serves only to modulate how the teacher scores student-sampled tokens without altering the student input or inference procedure. The teacher does not generate a separate response but exclusively evaluates the student's trajectory. The token-level distillation advantage is then computed as:
\begin{equation}
    \widehat{A}^{\mathrm{hint}}_t = \operatorname{sg}\!\left[ \log\pi_{T_d}\!\left(y_t \mid \mathcal{P}_T(x, z^*), y_{<t}\right) - \log\pi_{\theta}\!\left(y_t \mid \mathcal{P}_S(x), y_{<t}\right) \right],
    \label{eq:hinted-advantage}
\end{equation}
where $\widehat{A}^{\mathrm{hint}}_t$ is the distillation advantage for the $t$-th student-sampled token $y_t$, and $y_{<t}$ denotes its preceding context. $T_d$ is the frozen teacher for domain $d$, and $\pi_\theta$ is the student policy. The original GUI input is denoted by $x$, and $z^*$ is the correct action type. $\mathcal{P}_T(x, z^*)$ represents the teacher prompt augmented with the hint, whereas $\mathcal{P}_S(x)$ is the unchanged student prompt. Finally, $\operatorname{sg}[\cdot]$ stops gradients through the advantage computation. By incorporating these GUI-adaptive designs into our multi-teacher setting, MOPD ensures that the distillation signal is effectively concentrated on critical executable behaviors rather than reasoning traces, yielding significantly more robust fusion results across diverse GUI domains.

\section{Data Generation and Verification}
\label{sec:data_construction}

As shown in Figure~\ref{fig:venus20_overview}, mobile, web, and computer-use trajectories share a pipeline for capability discovery, task construction, rollout, and verification. Grounding and CAPTCHA use executable program states to supply target geometry and valid actions directly. Real-environment trajectories retain their task contracts and verification evidence, while synthesized records retain the program state that supplies their labels. We will detail data generation and verification as follows.

\subsection{General Generation Pipeline}\label{sec:general_generation}

Our data generation framework operates as a closed-loop iterative system designed to progressively expand task diversity and execution reliability. As illustrated in Figure~\ref{fig:venus20_overview}, the pipeline consists of three tightly coupled stages: (1) \textbf{Capability Catalog Construction}, which distills application knowledge from heterogeneous sources into a structured, dynamic registry; (2) \textbf{Task Construction}, which samples from this catalog to synthesize executable tasks with rigorous validity guarantees; and (3) \textbf{Trajectory Collection}, which executes these tasks in real environments and feeds observed outcomes back to update the catalog. This feedback mechanism ensures that task generation is continuously grounded in verified application capabilities and adapts to empirical failure patterns across collection rounds. We detail each stage below.

\paragraph{Capability catalogs.}
During the cold start phase, Deep Research aggregates application evidence from official documentation, help pages, user discussions, common workflows, and historical tasks. We organize this evidence into an application-specific capability catalog, where each entry records a function's signature, required objects and preconditions, compatible functions, and current task coverage. Crucially, this catalog is not static; once rollouts begin, it is dynamically updated with observed page states, UI control constraints, entity validations, and failure cases. These runtime observations directly govern the sampling distribution for subsequent rounds, ensuring the catalog reflects the true executable state space rather than merely documented specifications.

\paragraph{Task construction.}
Guided by the current state of the capability catalog, the generator synthesizes diverse task types, including single-capability, composite, query-based, batch, and scenario-based tasks. Each generated task is formalized as an executable contract that binds the natural language instruction to its application domain, capability tags, initial state, required resources, and expected outcome. To maintain dataset quality, a validity gate enforces strict constraints before execution, rejecting tasks that involve unsupported functions, ambiguous goals, missing dependencies, or unverifiable outcomes. This step ensures that only well-defined and executable tasks enter the trajectory collection stage.

\paragraph{Trajectory collection.}
The executor interacts with the declared environment through a screenshot-action loop, attempting to fulfill the contracted task. Both successful and failed executions yield structured feedback—including observed functions, invalid entities, unmet preconditions, and specific failure causes—which is written back to the capability catalog. This empirical feedback serves two purposes: it refines the accuracy of capability entries and informs the coverage-aware sampling strategy for the next generation round. Consequently, the pipeline autonomously shifts focus toward under-explored capabilities and previously failing scenarios, enabling systematic improvement over iterations.

\subsection{Domain-Specific Process}
\label{sec:domain_specific}

While the general pipeline (\S\ref{sec:general_generation}) provides a unified framework for catalog construction, task generation, and trajectory collection, each application domain presents unique environmental constraints and data requirements. Below, we detail how the core pipeline is instantiated and extended for four distinct domains. Crucially, all domain-specific outputs feed back into the shared capability catalog and adhere to the same validity gating and feedback mechanisms described above; we therefore focus exclusively on domain-adaptive components.


\paragraph{Web Navigation.}
Web collection requires scalable curation of high-quality interactive environments from the open internet. We construct a website pool by combining public browser-agent benchmarks with the Tranco ranking~\cite{pochat2018tranco}, then apply automated accessibility checks followed by Kimi~2.6 scoring on dynamicity, interactivity, content richness, and visual quality. This yields over 4{,}000 domains across 19 categories. Additionally, we seed the capability catalog with 45{,}000 tasks from InSTA-150k-v3~\cite{Trabucco2025InSTA}, prioritizing entries with rich success criteria annotations. Execution uses real Chrome sessions via a 15-action Playwright interface. Beyond the general validity gate, we apply domain-specific trajectory sanitization: step-level rules remove redundant waits, opposite-direction scrolling, and cyclic action patterns, while recoverable errors are preserved in otherwise successful trajectories to enrich failure-mode coverage in the catalog.

\paragraph{Computer Use.}
Desktop environments introduce stateful complexity absent in mobile and web domains, requiring deterministic environment setup and long-horizon task management. Each computer-use task is serialized as a TaskSpec containing a controlled desktop snapshot, setup operations, required files/services, source provenance, and an outcome evaluator. A materialization plan provisions the workspace and warmup state; fixture fingerprints deduplicate environments and detect hidden-answer leakage; preflight checks validate initial state integrity before execution enters the general rollout loop. For long-horizon tasks, we introduce hierarchical segmentation: tasks are split into subgoals with independent artifacts and completion checks. A verified exit state from one segment seeds recollection of subsequent segments, with accepted prefixes replayed to preserve context—extending the general trajectory collection mechanism to handle multi-session dependencies.

\paragraph{Synthetic GUI Grounding.}
Unlike the above domains that collect real-world trajectories, synthetic grounding generates pixel-aligned interaction targets to supplement point-grounding and bounding-box prediction capabilities. We synthesize executable HTML/CSS/JS interfaces from natural-language scenarios, optionally conditioned on personas or reference screenshots~\cite{yang2025scaling}. A single browser pass captures screenshot, DOM, element state, and geometry. Candidate targets are identified via DOM semantics and accessibility attributes, then validated through a nine-point hit test suite covering visibility, viewport clipping, scroll containment, occlusion, and painted-pixel checks; text-node refinement eliminates bounding boxes containing unrelated whitespace. Accepted targets are annotated with coordinates, visible text, ARIA attributes, and local context. Multiple intents may map to one target, while infeasible instructions serve as hard negatives. All records are exported directly for grounding-specific SFT and RL, bypassing the general task-construction stage. Examples are shown in Appendix Figure~\ref{fig:guisyn}.

\paragraph{Synthetic CAPTCHA.}
CAPTCHA data requires machine-verifiable ground truth that real-world collection cannot provide. We implement type-specific rule engines covering 70 CAPTCHA types, where each puzzle stores a latent state that deterministically defines its answer, target geometry, valid actions, and solution trace. A separate renderer composites puzzles into mobile panels or webpage contexts, mapping coordinates to the final canvas; puzzle validity is verified before rendering to ensure solvability. The resulting records include rendered observations, normalized geometry, reasoning supervision, and executable action sequences, exported for CAPTCHA-specialized SFT and RL training. This module operates independently of the general catalog but shares the same export format conventions. Examples are shown in Figure~\ref{fig:venusbench-captcha-overview}.

\subsection{Trajectory Verification}
\label{sec:verify}

Reliable reinforcement learning for GUI agents requires verification signals that are both semantically meaningful and temporally granular. Coarse binary success/failure labels derived from final-state checks are insufficient for open-ended tasks, where partial progress, reasoning quality, and task feasibility are equally critical to policy learning. To address this, we introduce a two-level trajectory verification framework that jointly assesses \textit{what} the agent accomplished and \textit{how} it arrived there. Trace-level verification evaluates the overall semantic alignment between the execution trace and the task objective, providing structured quality stratification for data curation. Sample-level verification complements this by assessing individual action correctness prior to execution, enabling fine-grained step-wise supervision and real-time intervention. Together, these two levels form a unified verification pipeline that supplies robust, multi-resolution reward signals while simultaneously feeding back into upstream task generation to improve data quality at scale.

\subsubsection{Trace-level Verification}
\label{sec:trace_verify}

Rule-based evaluators are effective for verifying file contents, configuration states, or other precisely readable final outcomes, but they struggle with open-ended GUI tasks where success is semantic rather than syntactic. To address this, we introduce \textbf{Semantic Guided Verification (SGV)}, a trajectory-level quality inspection strategy that complements rule-based checks by evaluating whether the agent's execution trace semantically satisfies the task objective based on  VLM as a Judge~\cite{sun2026osreward}. Crucially, SGV does not rely on the agent's self-declared success; trajectories that end in failure, timeout, or user escalation are equally eligible for verification.

The SGV pipeline proceeds in five stages: (1) extracting verifiable completion keypoints from the task objective; (2) segmenting the full trajectory into fixed-size windows and overlaying red markers on click (CLK) actions in screenshots to enhance visual grounding; (3) making parallel judgments on which keypoints are satisfied within each window, thereby accumulating process-level evidence; (4) applying hard rules to samples with unambiguous evidence, while routing ambiguous cases to a multimodal final judgment that integrates termination signals, behavioral statistics, per-window explanations, and the final screenshot; and (5) producing a structured report containing the conclusion, reasoning, per-keypoint status, and supporting evidence screenshots.

Each trajectory is classified into one of four categories: \textit{completed} (all critical objectives achieved), \textit{partial} (some objectives met or meaningful progress made toward completion), \textit{infeasible} (task objectively impossible due to external constraints, correctly identified by the agent), and \textit{failed} (no valid progress formed). For example, given the task ``Post specified content in a target group,'' SGV extracts three keypoints: entering the correct group, filling in the specified title and body, and confirming successful publication. A trajectory that completes the first two keypoints but exhausts its step budget before confirmation would be classified as \textit{partial} rather than \textit{failed}, since it demonstrates substantive progress distinguishable from zero-progress failures.

Importantly, SGV conclusions serve as a quality stratification mechanism for data curation rather than direct training labels or benchmark scores. Completed trajectories enter the high-quality candidate pool; partial trajectories are routed to a review queue for truncation or continuation-based repair; infeasible trajectories provide evidence for refining task feasibility rules; failed trajectories help disentangle deficiencies in task design, environment stability, and agent capability; and newly observed functionalities discovered during verification are fed back into the sub-capability pool to inform future task generation. This closed-loop feedback ensures that verification not only filters data but also actively improves the upstream task construction pipeline.

\subsubsection{Sample-level Verification}
\label{sec:sample_verify}

Complementing trace-level assessment, sample-level verification provides fine-grained supervision at each interaction step. Unlike post-hoc evaluation, we adopt an \textit{a priori} judgment paradigm that assesses action correctness based solely on information available \textit{before} execution—the current screenshot, declared action type and target, agent reasoning, and task goal. This design enables real-time intervention and avoids conflating action quality with downstream page transitions that may be influenced by external factors.

Each step is classified into one of four categories through a two-stage decision process: first checking reasoning-action consistency (whether the agent's stated intent matches its executed operation), then evaluating task alignment. 
\textbf{(1) Correct}: the action clearly advances the task and the agent demonstrates confident awareness of its correctness; 
\textbf{(2) Exploratory}: the agent exhibits explicit uncertainty but makes a reasonable attempt toward a plausible candidate path (e.g., probing an unfamiliar interface region), distinguishable from errors by its grounded rationale and from ineffective actions by its induced state change; 
\textbf{(3) Ineffective}: the action yields no substantive contribution, such as clicking non-interactive regions or entering short useless loops; 
\textbf{(4) Incorrect}: the action explicitly deviates from the task objective, continues after completion, or reflects reasoning-action inconsistency (e.g., perceptual hallucination or thought-action mismatch).

Step-level judgments are aggregated into a three-tier trajectory assessment. The first tier determines feasibility: infeasible tasks proceed to identification assessment (correctly reported $\rightarrow$ \textit{infeasible}; unrecognized $\rightarrow$ \textit{failed}). Feasible tasks undergo completion evaluation based on a critical distinction between \textit{generic operations} (app launching, tab switching, scrolling) and \textit{task-specific operations} (entering specific content, selecting particular targets). Only trajectories containing at least one valid task-specific operation qualify as \textit{partial}; those limited to generic operations are classified as \textit{failed}. This fine-grained aggregation ensures that partial credit reflects genuine task-grounded progress, providing a more reliable reward signal for reinforcement learning than coarse binary success/failure labels.

\section{Experiments}

\subsection{Experimental Setup}

\subsubsection{Implementation details}
During inference, we set the sampling temperature to 1.0 and use Qwen3.5’s default visual resolution configuration. For general agentic tasks, reasoning mode (think) is enabled by default, with the complete reasoning history retained and fed back into the context to preserve coherence across multi-step interactions. For GUI grounding tasks, we disable reasoning mode and set the temperature to 0 to enable direct and efficient localization. For CAPTCHA scenarios, our framework further supports multi-action parsing, allowing the model to generate a sequence of operations to complete verification challenges. Additional implementation details for individual benchmarks are provided in their respective experimental sections. For each column in the following tables, the best and second-best results are in \textbf{bold} and \underline{underlined}.




\subsubsection{Benchmarks}

\textbf{Mobile Use.}
We evaluate mobile-use capabilities on six benchmarks: {MobileGym}~\cite{wu2026mobilegym}, {VenusBench-Mobile}~\cite{venusbenchmobile2}, {AndroidWorld}~\cite{rawles2025androidworld}, {MobileWorld}~\cite{kong2026mobileworld}, {KnowUBench}~\cite{chen2026knowu}, and {MemGUI}~\cite{liu2026memgui}.
MobileGym provides a lightweight, browser-based simulation environment with full controllability for everyday mobile use. VenusBench-Mobile shifts the focus toward realistic, user-centric evaluation by assessing general-purpose mobile GUI agents on diverse, dynamic, and fine-grained everyday usage scenarios. AndroidWorld emphasizes realism and reproducibility through dynamically generated, parameterized tasks spanning real-world Android applications. Building on this line of work, MobileWorld introduces a more challenging setting centered on long-horizon, cross-application, user-aware, and hybrid-tool mobile use scenarios within a fully controlled and reproducible environment; in our report, however, we focus specifically on its GUI-based interaction setting. Beyond general mobile-use evaluation, KnowU-Bench targets personalized mobile assistance, testing agents in GUI-based scenarios that require user preference inference, multi-turn clarification, and calibrated proactive support. MemGUI-Bench, in contrast, centers on memory-intensive evaluation, examining whether mobile GUI agents can retain short-term information, accumulate long-term experience, and generalize across sessions and applications.

\textbf{Computer Use.}
We evaluate computer-use capabilities on {OSWorld-Verified} \cite{xie2024osworld}, {DeskCraft} \cite{wang2026deskcraft} and {OSWorld 2.0} \cite{yuan26osworld2}.
OSWorld-Verified evaluates agents on open-ended tasks in real computer environments, covering interactions with desktop and web applications, operating-system-level operations, and workflows spanning multiple applications.
DeskCraft complements OSWorld by targeting long-horizon professional desktop workflows and proactive human--agent collaboration. In addition to complex multi-step tasks, it explicitly evaluates agents' ability to interact with users during task execution, including seeking clarification under uncertainty, handling user interruptions, and incorporating post-execution feedback. OSWorld 2.0 further extends this evaluation to 108 long-horizon, real-world computer-use workflows spanning everyday and professional scenarios. Its tasks require agents to coordinate hundreds of actions across applications, reason over multiple information sources, adapt to dynamic environments and streaming updates.

\textbf{Web Navigation.}
We evaluate web-navigation capabilities on {WebVoyager} \cite{he2024webvoyager}, {Online-Mind2web} \cite{xue2025an}, {REAL} \cite{garg2026real} and {Odysseys} \cite{jang2026odysseys}, which provide complementary evaluation settings for real-world web interaction.
WebVoyager evaluates end-to-end multimodal web agents on tasks collected from popular real-world websites, emphasizing visual understanding and interaction with live web interfaces. 
Online-Mind2Web evaluates web agents on realistic tasks across live websites.
REAL complements these live-web benchmarks with deterministic, high-fidelity simulations of real websites, enabling controlled and reproducible evaluation of both information-retrieval and state-changing tasks. Odysseys focuses on long-horizon, multi-step workflows that span multiple websites on the live Internet, testing sustained navigation and cross-site reasoning.

\textbf{GUI Grounding.} We evaluate GUI grounding capabilities on {VenusBench-GD} \cite{zhou2025venusbench}, {ScreenSpot-Pro} \cite{li2024screenspot-pro}, {OSWorld-G-R} \cite{xie2025scalingcomputerusegroundinguser}, and {UI-Vision} \cite{nayak2025ui}, which provide complementary evaluation settings for fine-grained visual grounding in real-world interfaces. VenusBench-GD and ScreenSpot-Pro emphasize precise localization of interactive elements across diverse GUI environments, while OSWorld-G-R evaluates grounding in complex desktop tasks with functional interaction requirements. UI-Vision further assesses grounding under visually diverse and challenging interface layouts, providing a comprehensive evaluation of spatial understanding and action-oriented grounding capabilities.

\textbf{CAPTCHA Solving.}
We evaluate models on five complementary CAPTCHA benchmarks. We introduce
{VenusBench-CAPTCHA}, a deployment-oriented set of 219 screenshots spanning
eight interaction types, from OCR and ordered clicking to visual reasoning and
geometric manipulation. We additionally use the complete 1{,}050-example
{Spatial-CAPTCHA-Bench}~\cite{kharlamova2026spatialcaptcha}, a fixed,
category-balanced 1{,}000-example subset of {MCA-Bench}~\cite{wu2026mcabench},
319 examples from 15 selected {NextGen-CAPTCHAs}~\cite{liu2026nextgencaptchas}
task types, and 16 retained task types from {Open CaptchaWorld}~\cite{luo2025opencaptchaworld}.
Together, these benchmarks cover
spatial reasoning, visual recognition, logical reasoning, target grounding,
and executable GUI interaction. Detailed task coverage and selection criteria
are provided in Appendix~\ref{app:captcha-benchmarks}.

\textbf{GUI Agent Safety.} %
To assess safety in realistic desktop environments, we evaluate on {OSHarm}~\cite{kuntz2025osharm} and {OSBlind}~\cite{ding2026osblind}.
OSHarm benchmarks safety across three risk categories: deliberate user misuse, third-party prompt injection attacks, and model misbehavior (\emph{e.g.}, accidental costly mistakes).
OSBlind further evaluates susceptibility to safety blind spots under benign-looking instructions that lead to unintended harmful outcomes.
For both safety benchmarks, we report the Attack Success Rate (ASR), \emph{i.e.}, the fraction of tasks in which the agent carries out the harmful behavior.


\subsection{Main Results}
\begin{table*}[ht]
	\centering
	\footnotesize
	\setlength{\tabcolsep}{0pt}
	\begin{tabular*}{\textwidth}{@{\extracolsep{\fill}}lcccccc}
		\toprule
		\textbf{Models} & \textbf{MobileGym} & \textbf{VenusBench-Mobile} & \textbf{AndroidWorld} & \textbf{MobileWorld} & \textbf{KnowUBench} & \textbf{MemGUI} \\
		\midrule
		\rowcolor{gray!15}
		\multicolumn{7}{l}{\textit{General VLMs}} \\
		Qwen3.5-9B~\citep{qwen3.5} & 9.0* & 15.3* & 57.8 & 18.0(18.0)* & 33.3 & 6.2* \\
		Qwen3.6-27B~\citep{qwen3.6-27b} & 24.6* & 28.0* & 70.3 & 36.8(41.9)* & - & 25.7* \\
		Claude-Opus-4.6~\citep{anthropic2026opus46} & - & 36.5* & - & 44.5 & - & - \\
		Kimi-K2.6~\citep{kimik2.6} & 38.7* & 31.2* & - & 55.6 & - & 39.1 \\
		Kimi-K3~\citep{kimik3} & - & - & - & 74.4 & - & - \\
		Seed-2.0-Pro~\citep{seed2} & 52.0 & 20.1* & - & 63.2 & 51.6 & \underline{65.6}* \\
		Seed-2.1-Pro~\citep{seed21} & - & - & - & 73.2 & - & - \\
		GPT-5.6-Sol~\citep{openai2026gpt56} & - & - & - & 70.1 & - & - \\
		\midrule
		\rowcolor{gray!15}
		\multicolumn{7}{l}{\textit{GUI-specific Models}} \\
		UI-Venus-1.5-8B~\citep{team2026ui} & 18.4* & 16.1 & 73.7 & 22.2* & 26.0 & 3.9* \\
		UI-Venus-1.5-30B-A3B~\citep{team2026ui} & 21.5* & 21.5 & 77.6 & 17.1 & - & 10.9* \\
		GUI-Owl-1.5-32B-Instruct~\citep{guiowl1.5} & 20.3* & - & 69.8 & 43.9 & - & 10.9 \\
		MAI-UI-8B~\citep{zhou2025mai} & 21.5* & 12.7 & 70.7 & 27.5 & 26.0 & 17.2* \\
		Qwen-UI-Agent-27B~\citep{qwenuiagent} & - & - & - & \textbf{82.1(85.5)} & - & - \\
		\midrule
		\rowcolor{gray!15}
		\multicolumn{7}{l}{\textit{Ours}} \\
		\textbf{UI-Venus-2-9B}                                          & \underline{52.7} & \underline{46.5} & \underline{80.2} & 65.8(75.2) & \underline{56.5} & 62.6 \\
		\textbf{UI-Venus-2-27B}                                          & \textbf{60.5} & \textbf{48.7} & \textbf{84.0} & \underline{76.1(82.9)} & \textbf{59.7} & \textbf{70.3} \\
		\bottomrule
	\end{tabular*}
	\vspace{0.4em}
	\caption{
		Performance comparison on various mobile GUI benchmarks.
		VenusBench-Mobile reports success rate on its 149-task primary pool. For MobileWorld, we report GUI-only success rate on 117 tasks under the 50-step setting; values in parentheses, when available, use 100 steps.
		MemGUI reports Main Results pass@1. ``*'' denotes the baseline results evaluated or reproduced by us.
	}
	\label{tab:mobile-ues}
\end{table*}

In the experiments, we compare UI-Venus-2 models against a broad range of state-of-the-art baselines across two model categories:
\textbf{(1) General VLMs}: Qwen3.5-9B~\cite{qwen3.5}, Qwen3.6-27B~\cite{qwen3.6-27b}, Qwen3.7-Plus~\cite{qwen37plus}, Claude-Opus-4.6~\cite{anthropic2026opus46}, Kimi-K2.6~\cite{kimik2.6}, Kimi-K3~\cite{kimik3}, Seed-2.0-Pro~\cite{seed2}, Seed-2.1-Pro~\cite{seed21}, and GPT-5.6-Sol~\cite{openai2026gpt56}.
\textbf{(2) GUI-specific Models}: OpenCUA~\cite{wang2025opencuaopenfoundationscomputeruse}, GTA1~\cite{yang2025gta1guitesttimescaling}, GUI-Owl~\cite{ye2025mobile}, UI-TARS-1.5~\cite{ui-tars-15-seed}, UI-Venus~\cite{gu2025ui}, Holo2~\cite{hai2025holo2modelfamily}, Step-GUI~\cite{yan2025step}, MAI-UI~\cite{zhou2025mai}, UI-Venus-1.5~\cite{team2026ui}, and Qwen-UI-Agent~\cite{qwenuiagent}.

\subsubsection{Mobile Use}
\noindent \textbf{MobileGym.} On MobileGym, our models achieve the strongest performance among all compared methods, with UI-Venus-2-9B reaching 52.7\% and UI-Venus-2-27B further improving to 60.5\%. The 27B model surpasses the strongest baseline, Seed2.0 Pro (52.0\%), by 8.5\% points, while substantially outperforming all prior GUI-specific models, whose best result is 21.5\%. These results indicate that our models are particularly effective in controllable everyday mobile-use simulations, where robust perception and precise action grounding are critical.

\noindent \textbf{VenusBench-Mobile.} On VenusBench-Mobile, which emphasizes realistic, user-centric, and fine-grained evaluation, our models again achieve the best results, reaching 46.5\% and 48.7\% for the 9B and 27B variants, respectively. The strongest prior baseline is Opus 4.6 (36.5\%), followed by Kimi K2.6 (31.2\%) and Qwen3.6-27B (28.0\%). The substantial margin over both general-purpose and GUI-specific baselines suggests that our approach generalizes better to diverse and dynamic everyday mobile scenarios beyond app-centric benchmark settings.

\noindent \textbf{AndroidWorld.} On AndroidWorld, UI-Venus-2-9B and UI-Venus-2-27B achieve 80.2\% and 84.0\%, respectively, establishing the best results in the table. Notably, several strong baselines already perform well on this benchmark, including UI-Venus-1.5-30B-A3B (77.6\%), UI-Venus-1.5-8B (73.7\%), and MAI-UI-8B (70.7\%), suggesting that AndroidWorld is relatively mature and increasingly close to saturation. Despite this, our 27B model still improves over the previous best by 6.4\% points, demonstrating that our approach continues to yield meaningful gains even in a benchmark where headroom is already limited.

\noindent \textbf{MobileWorld.} MobileWorld is substantially more challenging due to its long-horizon and cross-application task structure. Under the GUI-only setting considered in our report, UI-Venus-2-9B achieves 65.8\% in the standard 50-step setting and 75.2\% in the 100-step setting, while UI-Venus-2-27B reaches 76.1\% and 82.9\%, respectively. At 50 steps, Qwen-UI-Agent-27B reports 82.1\%, while the strongest standalone general-model baselines include Kimi-K3 (74.4\%), Seed-2.1-Pro (73.2\%), and GPT-5.6-Sol (70.1\%). UI-Venus-2-27B therefore remains ahead of these selected general-model baselines, while trailing the strongest specialized agent.

\noindent \textbf{KnowUBench.} KnowUBench evaluates personalized mobile assistance, including user preference inference, multi-turn clarification, and calibrated proactive support. On this benchmark, UI-Venus-2-9B and UI-Venus-2-27B achieve 56.5\% and 59.7\%, respectively, both surpassing the strongest prior baseline, Seed2.0 Pro (51.6\%). The improvement is even larger relative to earlier GUI-specific models such as UI-Venus-1.5-8B (26.0\%) and MAI-UI-8B (26.0\%). These results suggest that our approach improves not only GUI interaction performance but also the user-aware capabilities required for personalized mobile assistance.

\noindent \textbf{MemGUI.} On MemGUI, which focuses on short-term retention, long-term experience accumulation, and cross-session learning, UI-Venus-2-27B achieves the best result of 70.3\%, while UI-Venus-2-9B reaches 62.6\%. The strongest baseline is Seed2.0 Pro (65.6\%), whereas other models remain substantially lower, including Kimi K2.6 (39.1\%), Qwen3.6-27B (25.7\%), and MAI-UI-8B (17.2\%). This substantial advantage indicates that our models are particularly strong in memory-intensive mobile GUI scenarios that require retaining and reusing information across actions, applications, and sessions.

\subsubsection{Computer Use}
\begin{table*}[ht]
	\begin{minipage}[t]{0.55\textwidth}
		\centering
		\footnotesize
		\setlength{\tabcolsep}{0pt}
		\begin{tabular*}{\textwidth}{@{\extracolsep{\fill}}lcc}
			\toprule
			\textbf{Models} & \textbf{OSWorld-Verified} & \textbf{DeskCraft} \\
			\midrule
			\rowcolor{gray!15}
			\multicolumn{3}{l}{\textit{General VLMs}} \\
			Claude-Opus-4.8~\citep{qwenuiagent} & \textbf{83.4} & - \\
			Qwen3.5-9B~\citep{qwen3.5} & 41.8 & 14.6$^*$ \\
			Qwen3.6-27B~\citep{qwen3.6-27b} & 62.0 & 28.7$^*$ \\
			Kimi-K2.6~\citep{kimik2.6} & 73.1 & 41.4$^*$ \\
			Seed-2.0-Pro~\citep{seed2} & 62.3 & 40.0$^*$ \\
			Seed-2.1-Pro~\citep{qwenuiagent} & 78.8 & - \\
			GPT-5.5~\citep{qwenuiagent} & 78.7 & - \\
			\midrule
			\rowcolor{gray!15}
			\multicolumn{3}{l}{\textit{GUI-specific Models}} \\
			GUI-Owl-1.5-32B-Instruct~\citep{guiowl1.5} & 56.5 & - \\
			Qwen-UI-Agent-27B~\citep{qwenuiagent} & 79.5 & - \\
			\midrule
			\rowcolor{gray!15}
			\multicolumn{3}{l}{\textit{Ours}} \\
			\textbf{UI-Venus-2-9B}      & 70.8 & \underline{48.0} \\
			\textbf{UI-Venus-2-27B}     & \underline{80.5} & \textbf{55.5} \\
			\bottomrule
		\end{tabular*}
	\end{minipage}%
	\hfill
	\begin{minipage}[t]{0.42\textwidth}
		\centering
		\footnotesize
		\setlength{\tabcolsep}{0pt}
		\begin{tabular*}{\textwidth}{@{\extracolsep{\fill}}lcc}
			\toprule
			\textbf{Models} & \textbf{Binary Acc.} & \textbf{Partial Score} \\
			\midrule
			\rowcolor{gray!15}
			\multicolumn{3}{l}{\textit{General VLMs}} \\
			GPT-5.5 & \textbf{13.0} & \textbf{46.7} \\
			Claude-Opus-4.7 & \underline{4.6} & \underline{20.3} \\
			Claude-Sonnet-4.6 (max) & \underline{4.6} & 20.0 \\
			Claude-Sonnet-4.6 (medium) & \underline{4.6} & 14.2 \\
			MiniMax-M3 & 1.9 & 8.2 \\
			Kimi-K2.6 & 1.9 & 7.1 \\
			Qwen3.5-9B & 0.0$^*$ & 2.5$^*$ \\
			Qwen3.6-27B & 0.0$^*$ & 3.8$^*$ \\
			Seed-2.0-Pro & 0.0$^*$ & 6.3$^*$ \\
			\midrule
			\rowcolor{gray!15}
			\multicolumn{3}{l}{\textit{Ours}} \\
			\textbf{UI-Venus-2-9B}      & 0.0 & 7.5 \\
			\textbf{UI-Venus-2-27B}     & 2.8 & 13.2 \\
			\bottomrule
		\end{tabular*}
	\end{minipage}
	\vspace{0.4em}
	\caption{
		Performance comparison on computer-use agent benchmarks: \textbf{OSWorld-Verified} and \textbf{DeskCraft} (left), and \textbf{OSWorld 2.0} (right) under the official 150-step budget with 108 tasks. Reported OSWorld-Verified baselines use the 361-task setting in their cited source and may use model-specific action scaffolds. For DeskCraft, we report an author-evaluated aggregate over the 538-task union of the Standard and Interactive splits, which differs from the benchmark's official split-level reporting. OSWorld 2.0 results report the official Binary Accuracy and Partial Score metrics; baselines are taken from the official leaderboard, possibly with model-specific tool settings, and the reasoning-effort setting is labeled in parentheses for models with multiple official entries. ``*'' indicates baseline results evaluated by us.
	}
	\label{tab:osworld}
\end{table*}

\noindent \textbf{OSWorld-Verified.}
On OSWorld-Verified, UI-Venus-2-27B achieves 80.5\%. This is below the source-reported Claude-Opus-4.8 result of 83.4\%, but above Qwen-UI-Agent-27B (79.5\%), Seed-2.1-Pro (78.8\%), and GPT-5.5 (78.7\%). UI-Venus-2-9B obtains 70.8\%. Because these reported systems use model-specific action scaffolds---including direct command-line actions for Qwen-UI-Agent---the comparison should be interpreted as a benchmark-level reference rather than a strictly controlled ablation.

\noindent \textbf{DeskCraft.}
On DeskCraft, UI-Venus-2-27B achieves the best overall performance of 55.5\%, outperforming the strongest baseline, Kimi-K2.6 (41.4\%), by a substantial 14.1 points and Seed-2.0-Pro (40.0\%) by 15.5 points. UI-Venus-2-9B ranks second overall with 48.0\%, exceeding Kimi-K2.6 by 6.6 points.

\noindent \textbf{OSWorld 2.0.}
On OSWorld 2.0, we evaluate our models under the official 150-step interaction budget and report both Binary Accuracy and Partial Score. UI-Venus-2-9B and UI-Venus-2-27B achieve Binary Accuracies of 0.0\% and 2.8\%, and Partial Scores of 7.5\% and 13.2\%, respectively. The 27B model has a Binary Accuracy of 2.8\%, compared with 13.0\% for GPT-5.5 and 4.6\% for Claude-4.x models. Its Partial Score of 13.2\% is higher than Kimi-K2.6 (7.1\%), MiniMax-M3 (8.2\%), and all evaluated Qwen3.5/3.6 and Seed-2.0-Pro variants, but remains below Claude-Opus-4.7 (20.3\%) and Claude-Sonnet-4.6 (20.0\%).

\subsubsection{Web Navigation}
\begin{table*}[t]
    \centering
    \footnotesize
    \setlength{\tabcolsep}{5pt}
    \renewcommand{\arraystretch}{1.1}

    \resizebox{0.85\textwidth}{!}{
    \begin{tabular}{lccccc}
        \toprule
        \multirow{2}{*}{\textbf{Models}}
        & \multirow{2}{*}{\textbf{WebVoyager}}
        & \multirow{2}{*}{\textbf{Online-Mind2Web}}
        & \multirow{2}{*}{\textbf{REAL}}
        & \multicolumn{2}{c}{\textbf{Odysseys}} \\
        \cmidrule(lr){5-6}
        &
        &
        &
        &
        \textbf{Avg.}
        & \textbf{Perfect}
        \\
        \midrule

        \rowcolor{gray!15}
        \multicolumn{6}{l}{\textit{General VLMs}} \\

        Qwen3.5-9B~\citep{qwen3.5}
        & 46.9$^{*}$
        & 27.3$^{*}$
        & 18.2$^{*}$
        & 42.6$^{*}$
        & 13.5$^{*}$ \\

        Qwen3.5-4B~\citep{qwen3.5}
        & --
        & --
        & --
        & 42.9
        & 10.7 \\

        Qwen3.6-27B~\citep{qwen3.6-27b}
        & 84.3$^{*}$
        & 55.3$^{*}$
        & 27.3$^{*}$
        & 39.5$^{*}$
        & 18.5$^{*}$ \\
        
        OpenAI Operator~\citep{openai2025operator}
        & 87.0
        & 61.3
        & --
        & --
        & -- \\

        GPT-5 (SoM)~\citep{awadallah2026fara}
        & 90.6
        & --
        & --
        & --
        & -- \\

        GPT-5.4~\citep{singh2025openai}
        & --
        & --
        & --
        & 55.4
        & 33.5 \\

    Seed-2.0-Pro~\citep{seed2}
        & 85.1$^{*}$
        & 68.5$^{*}$
        & 74.4$^{*}$
        & 60.2$^{*}$
        & 30.1$^{*}$ \\
    
        GLM-5V-Turbo~\citep{hong2026glm}
        & 88.5
        & --
        & --
        & --
        & -- \\

        Claude-Opus-4.6~\citep{anthropic2026opus46}
        & 88.0
        & --
        & --
        & 68.9
        & 44.5 \\

        Claude-Sonnet-4.6~\citep{anthropic2026sonnet46}
        & --
        & --
        & --
        & 49.8
        & 31.0 \\

        Kimi-K2.6~\citep{kimik2.6}
        & 76.8$^{*}$
        & --
        & 74.4$^{*}$
        & --
        & -- \\

        \midrule

        \rowcolor{gray!15}
        \multicolumn{6}{l}{\textit{GUI-specific Models}} \\

        UI-TARS-1.5~\citep{ui-tars-15-seed}
        & 84.8
        & \underline{75.8}
        & --
        & --
        & -- \\






        UI-Venus-1.5-30B-A3B~\citep{team2026ui}
        & 76.0
        & --
        & 38.0$^{*}$
        & --
        & -- \\



        GUI-Owl-1.5-32B-Thinking~\citep{guiowl1.5}
        & 82.1
        & --
        & 44.6$^{*}$
        & --
        & -- \\


        MolmoWeb-8B~\citep{gupta2026molmoweb}
        & 78.2
        & 35.3
        & --
        & --
        & -- \\


        Fara1.5-4B~\citep{awadallah2026fara}
        & 80.8
        & --
        & --
        & --
        & -- \\

        Fara1.5-9B~\citep{awadallah2026fara}
        & 86.6
        & 63.4
        & --
        & --
        & -- \\

        Fara1.5-27B~\citep{awadallah2026fara}
        & 89.3
        & 72.3
        & --
        & --
        & -- \\

        \midrule

        \rowcolor{gray!15}
        \multicolumn{6}{l}{\textit{Ours}} \\

        \textbf{UI-Venus-2-9B}
        & \underline{90.8}
        & 74.0
        & \underline{76.9}
        & \underline{77.3}
        & \underline{62.0} \\

        \textbf{UI-Venus-2-27B}
        & \textbf{93.4}
        & \textbf{78.3}
        & \textbf{80.2}
        & \textbf{80.4}
        & \textbf{66.3} \\

        \bottomrule
    \end{tabular}
    }

    \caption{
    Performance comparison on four live-web benchmarks:
    \textbf{WebVoyager}, \textbf{Online-Mind2Web}, \textbf{REAL}, and \textbf{Odysseys}.
    The Fara1.5 and GPT-5 (SoM) WebVoyager entries use the refreshed 595-task, 100-step robust protocol and are averaged over three runs; live-site states may vary by evaluation date.
    For Odysseys, we report both the averaged rubric score (Avg.) and
    the perfect rubric score (Perfect). 
  "*" indicates our reproduced results.
    }
    \label{tab:web_navigation}
\end{table*}

\noindent \textbf{WebVoyager.}
WebVoyager is a live-web benchmark for evaluating end-to-end web navigation agents. The original benchmark contains 643 tasks across 15 popular real-world websites, covering information seeking, search, and interactive navigation.
Following the evaluation protocol of \cite{fara7b2025}, we use a filtered and refreshed version of WebVoyager: approximately 48 tasks that are no longer feasible are removed, while around 50 time-sensitive tasks are updated with valid dates to remain executable in the current web environment.
Following the official evaluation protocol, task completion is evaluated using an automatic GPT-4o judge based on the agent's interaction trajectory and final response.  As shown in Table~\ref{tab:web_navigation}, UI-Venus-2-27B achieves the best performance of 93.4\%. It outperforms the strongest GUI-based baseline, Fara1.5-27B, by 4.1 points. Compared with GPT-5 (SoM), UI-Venus-2-27B also achieves a 2.8-point improvement. Moreover, UI-Venus-2-9B reaches 90.8\%, retaining strong performance at a substantially smaller model scale.

\noindent \textbf{Online-Mind2Web.}
Online-Mind2Web is a live-web benchmark designed to evaluate web agents in realistic online environments. It consists of 300 diverse tasks across 136 real-world websites, with task completion assessed by an LLM-based automatic evaluator. UI-Venus-2-27B achieves the best performance of 78.3\%. UI-Venus-2-9B model also achieves 74.0\%, outperforming Fara1.5-27B despite its substantially smaller scale.

\noindent \textbf{REAL.}
REAL evaluates web agents on 112 tasks across 11 high-fidelity replicas of real-world web applications. Unlike live-web benchmarks, REAL provides deterministic and reproducible environments while preserving realistic interfaces and state-changing interactions. The tasks span information retrieval, action execution, and their combinations, with action completion verified through programmatic state assertions. UI-Venus-2-27B achieves a new state-of-the-art score of 80.2\%, outperforming the strongest baseline of 74.4\% by 5.8 points. The 9B model also reaches 76.9\%, exceeding the previous best by 2.5 points,

\noindent \textbf{Odysseys.}
Odysseys is a long-horizon, cross-site benchmark built on the live web, containing 200 tasks derived from realistic human browsing scenarios. Following the official evaluation protocol, we use \texttt{gemini-3.1-flash-lite-preview} as the judge to independently evaluate each rubric item based on the agent's interaction trajectory. We report two rubric-based metrics: \textbf{Averaged}, which measures the average fraction of satisfied rubric items, and \textbf{Perfect}, which counts a task as successful only when all rubric items are satisfied. UI-Venus-2-27B achieves the best results on both metrics, reaching 80.4 in averaged rubric score and 66.3 in perfect rubric score, outperforming the strongest baseline by 11.5 and 21.8 points, respectively. The 9B model also attains 77.3 and 62.0, substantially surpassing existing models and demonstrating strong capability.

\subsubsection{GUI Grounding}

\noindent \textbf{VenusBench-GD.} VenusBench-GD is a comprehensive grounding benchmark spanning web, desktop, and mobile UIs, covering both basic localization and reasoning-intensive cases, including \emph{refusal grounding} for infeasible instructions. As shown in Table~\ref{tab:grounding_benchmarks}, UI-Venus-2-27B achieves a new state-of-the-art accuracy of 80.1\%, improving over UI-Venus-1.5-30B-A3B by 5.1 points. Notably, the 9B model also reaches 77.1\%, demonstrating strong grounding performance at a substantially smaller scale.

\noindent \textbf{ScreenSpot-Pro.} ScreenSpot-Pro targets high-resolution professional software interfaces, including CAD, development, creative, and office applications, where dense layouts and small interactive elements require precise fine-grained grounding. UI-Venus-2-27B achieves 74.1\%, ranking second only to Qwen-UI-Agent-27B (76.6\%) and improving by 4.5 points over UI-Venus-1.5-30B-A3B. Notably, it also surpasses strong closed-source flagship foundation models such as Qwen 3.7 Plus (68.9\%) and Seed 2.1 Pro (65.3\%). The 9B model also attains 73.0\%, substantially narrowing the gap to the 27B model.

\begin{table*}[t]
	\centering
	\footnotesize
	\setlength{\tabcolsep}{6pt}
	\begin{tabular}{l *{4}{c}}
		\toprule
		\multirow{2}{*}{\textbf{Models}} &
		\multicolumn{4}{c}{\textbf{Grounding Benchmarks}} \\
		\cmidrule(lr){2-5}
		& \textbf{VenusBench-GD} & \textbf{ScreenSpot-Pro} & \textbf{OSworld-G-R} & \textbf{UI-Vision} \\
		\midrule
		\rowcolor{gray!15}
		\multicolumn{5}{l}{\textit{General VLMs}} \\
        Qwen 3.7 Plus~\citep{qwen37plus}  & 75.2* & 68.9 & 78.2 & \underline{68.0} \\
		Seed 2.1 Pro~\citep{seed21}                       & 73.9* & 65.3 & 78.0 & 62.0 \\
        Kimi-K2.6~\citep{kimik2.6} & 73.1* & 52.0* & 69.7* & 51.7*\\
        Qwen3.6-27B~\citep{qwen3.6-27b} & 67.7* & 65.2* & 76.9* & 58.3*\\
		\midrule
		\rowcolor{gray!15}
		\multicolumn{5}{l}{\textit{GUI-specific Models}} \\
		UI-Venus-Ground-72B~\citep{gu2025ui}                 & {70.2} & {61.9} & 69.5 & {36.8} \\
		Holo2-30B-A3B~\citep{hai2025holo2modelfamily}                    & 59.5* & 66.1 & 76.1 & 40.9* \\
		Step-GUI-4B~\citep{yan2025step}                      & 54.6* & 60.0 & 66.9 & 30.0* \\
		MAI-UI-8B~\citep{zhou2025mai}                        & 65.2* & 65.8 & 68.6 & 40.7 \\
		MAI-UI-32B~\citep{zhou2025mai}                       & - & {67.9} & 73.9 & 47.1 \\
		UI-Venus-1.5-30B-A3B~\citep{team2026ui}             & 75.0 & 69.6 & 76.4 & 54.7 \\
        Qwen-UI-Agent-27B~\citep{qwenuiagent}& - & \textbf{76.6} & 78.5 & \textbf{70.0} \\
        \midrule
        \rowcolor{gray!15}
		\multicolumn{5}{l}{\textit{Ours}} \\
		\textbf{UI-Venus-2-9B}         & \underline{77.1} & 73.0 & \underline{78.5} & 53.2 \\
        \textbf{UI-Venus-2-27B}        & \textbf{80.1} & \underline{74.1} & \textbf{79.1} & 66.9 \\
    \bottomrule
	\end{tabular}
	\vspace{0.4em}
	\caption{Performance comparison on various \textbf{Grounding Benchmarks}. VenusBench-GD reports English-instruction micro-average point-in-box accuracy. 
    ``*'' indicates baselines evaluated or reproduced by us.}
	\label{tab:grounding_benchmarks}
\end{table*}

\noindent \textbf{OSWorld-G-R.} OSWorld-G-R evaluates grounding in complex desktop environments, requiring accurate localization of functionally relevant UI elements. UI-Venus-2-27B achieves 79.1\%, the highest score among GUI-specific models, while UI-Venus-2-9B reaches 78.5\%, demonstrating robust grounding capability across different model scales.

\noindent \textbf{UI-Vision.} UI-Vision evaluates grounding under diverse and visually challenging interface layouts. UI-Venus-2-27B achieves 66.9\%, second only to Qwen-UI-Agent-27B (70.0\%) yet substantially improving over UI-Venus-1.5-30B-A3B by 12.2 points. Together, these results demonstrate the strong and consistent grounding capability of UI-Venus-2 across heterogeneous GUI environments.

\subsubsection{CAPTCHA Solving}

\noindent \textbf{VenusBench-CAPTCHA.}
On the deployment-oriented VenusBench-CAPTCHA, models must convert CAPTCHA
understanding into complete executable solutions, including ordered clicks,
text entry, and geometrically accurate drags. Table~\ref{tab:venusbench-captcha-results}
shows that UI-Venus-2-27B and UI-Venus-2-9B obtain 79.9\% and 78.1\% Pass@1,
respectively, compared with 53.0\% for the strongest general-purpose model,
Qwen3.6-27B. The resulting gains of 26.9 and 25.1 percentage points indicate
that both variants are effective not only at recognizing CAPTCHA content but
also at translating their decisions into valid interface actions. Moreover,
the two UI-Venus-2 models differ by only 1.8 points, suggesting that the 9B
variant preserves most of the capability required by this end-to-end
perception--reasoning--action setting.

\noindent \textbf{MCA-Bench.}
The category-balanced MCA-Bench subset examines whether performance transfers
across a broad task distribution rather than concentrating on a small number
of common CAPTCHA formats. Each of its 20 categories contributes 50 examples,
giving equal weight to static recognition, target localization, interactive
manipulation, and logic-based question answering. On this diverse mixture,
UI-Venus-2-27B achieves 79.6\% Pass@1 and UI-Venus-2-9B achieves 75.7\%, while
the strongest general-purpose VLM, Qwen3.6-27B, reaches 51.7\%. Thus, even the
9B model improves over the best general-purpose baseline by 24.0 percentage
points, with the 27B model extending the margin to 27.9 points. Because no
single category dominates the aggregate score, these gains provide evidence
that UI-Venus-2 generalizes across substantially different recognition,
reasoning, and interaction requirements.

\begin{table*}[t]
  \centering
  \footnotesize
  \setlength{\tabcolsep}{3pt}
  \begin{tabular*}{\textwidth}{@{\extracolsep{\fill}}lccccc}
    \toprule
    \textbf{Models}
    & \textbf{\shortstack{VenusBench-\\CAPTCHA}}
    & \textbf{MCA-Bench}
    & \textbf{\shortstack{Spatial-\\CAPTCHA-Bench}}
    & \textbf{\shortstack{NextGen-\\CAPTCHAs}}
    & \textbf{\shortstack{Open\\CaptchaWorld}} \\
    \midrule
    \rowcolor{gray!15}
    \multicolumn{6}{l}{\textit{General VLMs}} \\
    Qwen3.5-9B~\citep{qwen3.5}           & 28.3 & 30.4 &  4.9 &  2.8 & 36.4 \\
    Qwen3.6-27B~\citep{qwen3.6-27b}      & 53.0 & 51.7 & 31.0 & 14.1 & 47.7 \\
    Seed-2.0-Pro~\citep{seed2}    & 47.9 & 36.5 & \underline{43.8}
                                           & 20.4 & \underline{55.6} \\
    Kimi-K2.6~\citep{kimik2.6}           & 39.7 & 38.7 & 24.8 &  7.2 & 47.8 \\
    Claude-Opus-4.6~\citep{anthropic2026opus46}      & 16.0 & 25.9 &  9.5 &  2.8 & 23.3 \\
    \midrule
    \rowcolor{gray!15}
    \multicolumn{6}{l}{\textit{Ours}} \\
    \textbf{UI-Venus-2-9B}
      & \underline{78.1} & \underline{75.7} & 42.8 & \underline{47.6}
      & 50.7 \\
    \textbf{UI-Venus-2-27B}
      & \textbf{79.9} & \textbf{79.6} & \textbf{48.6}
      & \textbf{54.5} & \textbf{56.3} \\
    \bottomrule
  \end{tabular*}

  \vspace{0.4em}
  \caption{\textbf{Performance Comparison across CAPTCHA Benchmarks}.
  All results are Pass@1 percentages, and higher is better. We evaluate on
  VenusBench-CAPTCHA and four public benchmarks:
  MCA-Bench~\cite{wu2026mcabench},
  Spatial-CAPTCHA-Bench~\cite{kharlamova2026spatialcaptcha},
  NextGen-CAPTCHAs~\cite{liu2026nextgencaptchas}, and
  Open CaptchaWorld~\cite{luo2025opencaptchaworld}. We use 1{,}000 sampled
  MCA-Bench examples, 15 NextGen-CAPTCHAs task types, and 16 Open CaptchaWorld
  task types; see the appendix for selection details.}
  \label{tab:venusbench-captcha-results}
\end{table*}

\noindent \textbf{Spatial-CAPTCHA-Bench.}
Spatial-CAPTCHA-Bench presents a markedly more competitive setting. Over the
complete set of 1{,}050 instances, UI-Venus-2-27B records 48.6\% Pass@1, a
4.8-point improvement over Seed-2.0-Pro at 43.8\%. UI-Venus-2-9B reaches
42.8\%, only 1.0 point below Seed-2.0-Pro but 37.9 points above the
similarly sized Qwen3.5-9B. This same-scale comparison reveals a substantial
advantage for UI-Venus-2 beyond what can be attributed to parameter count
alone. At the same time, scaling UI-Venus-2 from 9B to 27B yields a 5.8-point
gain, and even the best result remains below 50\%. Together, these observations
indicate that reference-frame changes, perspective taking, and multi-step
spatial transformations remain important sources of error for all evaluated
models, while additional model capacity provides a measurable benefit.

\noindent \textbf{NextGen-CAPTCHAs.}
The clearest separation from general-purpose VLMs occurs on
NextGen-CAPTCHAs. Across 319 examples from 15 retained task types,
UI-Venus-2-27B achieves 54.5\% Pass@1 and UI-Venus-2-9B achieves 47.6\%, whereas
the strongest general-purpose baseline, Seed-2.0-Pro, reaches only
20.4\%. The corresponding improvements of 34.1 and 27.2 percentage points are
the largest observed in our evaluation, and the 9B model alone more than
doubles the baseline score. These results highlight UI-Venus-2's strength on
CAPTCHAs that combine multiple reasoning primitives, such as counting,
rotation, occlusion, viewpoint changes, and path analysis. The 6.9-point gap
between the two UI-Venus-2 variants is also the largest model-scale gap among
the five benchmarks, suggesting that additional capacity is particularly
useful when several visual transformations or intermediate deductions must be
integrated into a single prediction.

\noindent \textbf{Open CaptchaWorld.}
Open CaptchaWorld produces the tightest comparison in
Table~\ref{tab:venusbench-captcha-results}.
Across its 16 retained task types, UI-Venus-2-27B reaches 56.3\% Pass@1, only
0.7 percentage points above Seed-2.0-Pro at 55.6\%. UI-Venus-2-9B ranks
third at 50.7\%: it trails Seed-2.0-Pro by 4.9 points but remains 2.9
points ahead of the next-best general-purpose model, Kimi-K2.6. Unlike the
large margins observed on VenusBench-CAPTCHA, MCA-Bench, and
NextGen-CAPTCHAs, this near parity indicates that strong general-purpose VLMs
already handle a considerable portion of the recognition, matching, counting,
and path-finding skills represented in Open CaptchaWorld. The 27B model's
leading result nevertheless demonstrates broad coverage, while the modest
margin and the remaining gap for the 9B variant expose clear headroom on
heterogeneous CAPTCHA mechanisms.

\subsubsection{GUI Agent Safety}
\begin{table}[t]
\centering
\small
\begin{tabular}{lcc}
\toprule
\textbf{Model} & \textbf{OSHarm} (ASR $\downarrow$) & \textbf{OSBlind} (ASR $\downarrow$) \\
\midrule
\rowcolor{gray!15}
\multicolumn{3}{l}{\textit{General VLMs}} \\
Qwen3.5-9B~\citep{qwen3.5} & 25.3 & 79.4 \\
Qwen3.5-27B~\citep{qwen3.5} & 18.0 & 89.3 \\
Kimi K2.6~\citep{kimik2.6} & 32.0 & 93.6 \\
\midrule
\rowcolor{gray!15}
\multicolumn{3}{l}{\textit{GUI-specific Models}} \\
EvoCUA-8B~\citep{xue2026evocua} & 39.3 & 85.3 \\
EvoCUA-32B~\citep{xue2026evocua} & 33.3 & 90.7 \\
UI-TARS-1.5~\citep{ui-tars-15-seed} & 36.0 & 83.3 \\
ScaleCUA~\citep{lv2026scalecuascalingcomputeruse} & 25.3 & 84.7 \\
\midrule
\rowcolor{gray!15}
\multicolumn{3}{l}{\textit{Ours}} \\
\textbf{UI-Venus-2-9B} & \textbf{11.3} & \underline{48.8} \\
\textbf{UI-Venus-2-27B} & \underline{15.3} & \textbf{47.9} \\
\bottomrule
\end{tabular}
\vspace{0.4em}
\caption{Safety evaluation on OSHARM and OS-BLIND benchmarks. ASR denotes Attack Success Rate (\%, lower is better).}
\label{tab:safety_eval}
\end{table}

\noindent \textbf{OS-Safety.}
We evaluate the agent's safety on two complementary threat settings and report the Attack Success Rate (ASR, lower is better).
On OSHarm, which targets explicit threats such as deliberate misuse and prompt injection, UI-Venus-2-9B achieves the lowest ASR of 11.3\% and UI-Venus-2-27B follows at 15.3\%, both outperforming the strongest baseline, Qwen3.5-27B (18.0\%).
On OSBlind, where instructions are entirely benign and harm arises from the execution context, all baselines remain highly vulnerable, with ASR ranging from 79.4\% to 93.6\%.
In contrast, UI-Venus-2-9B and UI-Venus-2-27B cut the ASR to 48.8\% and 47.9\%, reducing the attack success rate by up to nearly half relative to their base counterparts (79.4\% and 89.3\%, respectively).
These results highlight that our models not only resist explicit attacks but also reliably recognize safety blind spots during benign-looking execution, achieving a favorable balance between safety and capability.

\section{Related Works}

As digital tasks increasingly span mobile devices, web browsers, and desktop operating systems, GUI research is moving from isolated operations toward general-purpose interaction. Practical agents must therefore integrate perception, grounding, reasoning, planning, and interaction capabilities across domains.

\textbf{Mobile Use.}
Earlier mobile-use studies primarily relied on static screenshots, recorded trajectories, or fixed benchmarks, which provide reproducible measurements but only partially reflect practical interaction. Recent work increasingly evaluates agents in real applications and dynamic device environments, where interface states, application versions, languages, accounts, and interruptions can affect execution \cite{qin2025uitarspioneeringautomatedgui,zeng2025uitron,sun2025guixploreempoweringgeneralizablegui,yang2024ariauivisualgroundinggui,sun2025osgenesisautomatingguiagent,qiu2026unified,wang2025ui,liu2024autoglm,yan2025step,cao2026xiaomi,team2026hymobileagent}. This shift places greater demands on state tracking, error recovery, and long-horizon decision making.

\textbf{Computer Use.}
Desktop systems and professional software often contain denser interfaces and richer action spaces than mobile applications. Tasks may require keyboard shortcuts, file manipulation, text selection, dragging, window management, and coordination across multiple applications \cite{li2025dart,wang2025opencuaopenfoundationscomputeruse,yang2026ultracuafoundationmodelcomputer,xue2026evocua,qwenuiagent,guiowl1.5}. Their typically longer action sequences require comprehensive state understanding, planning, memory, progress assessment, recovery, and safe execution.

\textbf{Web Navigation.}
The live web is inherently uncertain: network failures may interrupt execution, content can update in real time, and authentication, pop-ups, advertisements, or regional differences can alter the interaction path \cite{gupta2026molmoweb,fara7b2025,awadallah2026fara, browseragent, wuying, webagent, browser_use2024}. Cross-site tasks add further uncertainty because different websites evolve independently. Successful navigation therefore requires continuous reassessment and adaptation to changes not observed during training.

\textbf{GUI Grounding.}
Grounding becomes more demanding on high-resolution screens and professional software, where precise localization and domain knowledge are needed to distinguish dense or visually similar controls \cite{lu2025uir1enhancingefficientaction,luo2025guir1generalistr1style,liu2025infiguir1advancingmultimodalgui,zhou2025guig1understandingr1zeroliketraining,yuan2025enhancingvisualgroundinggui,tang2025lpoaccurateguiagent,guig2,zhang2026omegausebuildinggeneralpurposegui,zhou2025mai,qin2025uitarspioneeringautomatedgui,ui-tars-15-seed,yao2026autofocus,yan2025step,hai2025holo2modelfamily,yang2025gta1guitesttimescaling,zhang2025mvp,vista}. Targets may also be ambiguous, differ from their descriptions, or be absent from the current screen. Agents must therefore combine visual, functional, and contextual evidence while avoiding incorrect actions when no valid target exists.

\textbf{CAPTCHA Solving.}
Modern CAPTCHAs extend beyond distorted-text recognition to ordered clicking, rotation, sliders, dragging, and spatial reasoning. Vision-language models can address several formats through shared visual understanding and GUI actions \cite{seed2,seed21,qwen38,kimik2.6,anthropic2025claude4}, although benchmark performance may not reflect deployed interaction patterns. Reliable solving requires fine-grained perception, spatial reasoning, precise grounding, and appropriate action generation.

\textbf{GUI Agent Safety.}
Real-world deployment is safety-critical for agents, especially computer-use agents whose actions can have costly consequences for users.
Early evaluations primarily examine whether agents can avoid harmful actions in explicitly risky or adversarial settings \cite{andriushchenko2024agentharm,zhan2024injecagent,kuntz2025osharm}.
Recent work reveals a subtler blind spot: harmful outcomes can also arise from seemingly benign instructions when latent risks are embedded in the environment \cite{ding2026osblind,yin2024safeagentbench}.
A trustworthy agent should safeguard users regardless of how the underlying risk arises, \emph{e.g.}, from erroneous user instructions, environmental hazards, or agent misalignment.

\section{Conclusion}

In this work, we presented UI-Venus-2, a GUI agent with state-of-the-art performance achieved through the joint scaling of multilingual environments, desktop computer-use capabilities, and fine-grained verification mechanisms. Beyond benchmark results, we pay particular attention to generalization abilities across diverse platforms such as mobile, web, and desktop operating systems to better serve realistic human everyday use. Looking forward, we hope UI-Venus-2 serves as a reproducible baseline and a collaborative platform for the community to further advance safe, reliable, and truly general-purpose GUI Agents.


\section{Contributions}
All contributors of UI-Venus-2 are listed in alphabetical order by their last names.


\begin{multicols}{4} 
\noindent
Zhuohan~Cai \\
Haoxing~Chen \\
Jiaxuan~Chen \\
Weizhi~Chen \\
Changlong~Gao \\
Zhangxuan~Gu \\
Yuan~Guo \\
Yusong~Hu \\
Jianrong~Jiang \\
Jianguo~Li \\
Runze~Li \\
Jinzhen~Lin \\
Zhenyu~Ma \\
Changhua~Meng$^{\dag}$ \\
Han~Peng \\
Xinyu~Qiu \\
Shuheng~Shen$^{\dag}$ \\
Zhongyi~Shui \\
Weiqiang~Wang \\
Ming~Wen \\
Zhuoer~Xu \\
Hang~Yan \\
Kaiwen~Yang \\
Ruilin~Yao \\
Nanjun~Yu \\
Zhengwen~Zeng \\
Lianrui~Zhang \\
Yunzhu~Zhang \\
Zhe~Zhao \\
Beitong~Zhou
\end{multicols}





\footnote{$^{\dag}$Corresponding Authors: Shuheng~Shen(shuheng.ssh@antgroup.com), Changhua~Meng(changhua.mch@antgroup.com).}

\newpage
\bibliographystyle{assets/plainnat}
\bibliography{main}

\newpage

\begin{appendix}

	\section{Action Space and Prompt Templates}

\subsection{Action Space}
\begin{table*}[ht]
\centering
\renewcommand{\arraystretch}{1.1} 
\scalebox{0.8}{
\begin{tabular}{ll}
\toprule
\textbf{Action} &\textbf{Definition}\\
\midrule
\multicolumn{2}{l}{\textit{\textbf{Shared Actions (All Platforms)}}}\\
\midrule
    Click(point=(x, y)) & Click at coordinates (x, y).\\
    Drag(start=(x1, y1), end=(x2, y2)) & Drag from (x1, y1) to (x2, y2).\\
    Swipe(start=(x1, y1), end=(x2, y2)) & Scroll by swiping from (x1, y1) to (x2, y2).\\
    DoubleClick(point=(x, y)) & Perform a double-click/tap at coordinates (x, y).\\
    LongPress(point=(x, y)) & Long press at coordinates (x, y) to trigger extra options.\\
    Type(content='') & Type the specified content.\\
    Wait() & Wait for loading.\\
    CallUser(content='') & Request user takeover or additional information.\\
    Finished(content='') & Mark the task as completed, with optional information.\\
\midrule
\multicolumn{2}{l}{\textit{\textbf{Mobile}}\footnote{Coordinates are normalized to $[0,999]$.}}\\
\midrule
    PressBack() & Press the `back' button.\\
    PressHome() & Press the `home' button.\\
    PressEnter() & Press the `enter' button.\\
    PressRecent() & Press the `recent' button.\\
    LaunchApp(app='') & Launch the specified app.\\
    GetScreenshot() & Take a screenshot and save it to the photo album.\\
    Answer(content='') & Answer the user's questions as requested.\\
\midrule
\multicolumn{2}{l}{\textit{\textbf{Desktop}}\footnote{Coordinates are normalized to $[0,999]$.}}\\
\midrule
    RightClick(box=(x, y)) & Right-click at (x, y) to open context menus.\\
    Hotkey(keys=[`ctrl', `c']) & Press a keyboard shortcut, e.g., Ctrl+C for copy.\\
\midrule
\multicolumn{2}{l}{\textit{\textbf{Web}}\quad\footnote{Coordinates are normalized to $[0,999]$.}}\\
\midrule
    Scroll(point=(x, y), direction=`up/down/left/right') & Scroll at (x, y) in the specified direction.\\
    Launch(url='') & Launch the target URL.\\
    GetUrl() & Get the URL of the current browser tab.\\
    TakeNote(content='') & Record important information from the screenshot.\\
    Hover(point=(x, y)) & Move the mouse cursor to coordinates (x, y) without clicking. \\
    Hotkey(keys=(`ctrl', `c')) & Press combination keys (up to 3).\\
    SelectOption(index=3) & Choose an option from a native HTML select element.\\
    PressBack() & Return to the previous page.\\
    PressHome() & Return to the browser home page.\\
    PressEnter() & Perform an `enter' key action.\\
\bottomrule
\end{tabular}
}
\caption{All actions and their definitions used in \textbf{UI-Venus-2}. We unify the action space and map all the actions in the existing open-source dataset to this space. }
\label{tab:action_space}
\end{table*}

\subsection{Grounding}

\begin{tcolorbox}[
		title=Grounding Prompt]\label{grounding_prompt}
	Output the center point of the position corresponding to the following instruction: \texttt{\color{red} \{problem\}}. The output should just be the coordinates of a point, in the format [x,y]. Additionally, if the task is infeasible (e.g., the task is not related to the image), the output should be [-1,-1].
\end{tcolorbox}

\subsection{Mobile}

\begin{tcolorbox}[
    title=Mobile Prompt,
    breakable
]\label{mobile_prompt}

\textbf{You are a GUI Agent.}

Your role is to analyze the user's task, provide clear and accurate answers to their questions, and execute the task with precise actions.

\medskip

\#\#\# Available Actions

You may execute one of the following functions:

\begin{itemize}
    \item \texttt{Click(box=(x1, y1))}

    Perform a tap action at the specified screen coordinate. Valid coordinates range from the top-left corner $(0,0)$ to the bottom-right corner $(999,999)$.

    \item \texttt{Drag(start=(x1, y1), end=(x2, y2))}

    Perform a drag action by long-pressing at the start coordinate for a few seconds and then dragging to the end coordinate. This is typically used for adjusting app layouts, moving sliders, solving slider captchas, etc. Valid coordinates range from $(0,0)$ to $(999,999)$.

    \item \texttt{Swipe(start=(x1, y1), end=(x2, y2))}

    Perform a swipe action by dragging from the start coordinate to the end coordinate. This is typically used for scrolling to find content, switching tabs, pulling down the notification shade, etc. Valid coordinates range from $(0,0)$ to $(999,999)$.

    \item \texttt{DoubleClick(box=(x1, y1))}

    Perform a double tap action at the specified screen coordinate. Valid coordinates range from $(0,0)$ to $(999,999)$.

    \item \texttt{LongPress(box=(x1, y1))}

    Perform a long-press action at the specified screen coordinate for a certain duration. This can be used to trigger additional options, such as copy, forward, delete, etc. Valid coordinates range from $(0,0)$ to $(999,999)$.

    \item \texttt{Type(content='')}

    Enter the specified text into the currently active input field.

    \item \texttt{LaunchApp(app='')}

    Launch the target app. Use this action when the target app is not currently visible on the screen.

    \item \texttt{Wait()}

    Wait for the current page, animation, or content to finish loading.

    \item \texttt{CallUser(content='')}

    Request user takeover or additional information when needed, for example, when there are multiple on-screen options that satisfy the requirement.

    \item \texttt{GetScreenshot()}

    Take a screenshot and save it to the device's photo album.

    \item \texttt{PressBack()}

    Return to the previous screen.

    \item \texttt{PressHome()}

    Return to the system home screen.

    \item \texttt{PressEnter()}

    Perform an Enter key action.

    \item \texttt{PressRecent()}

    Open the system recent apps screen.

    \item \texttt{Answer(content='')}

    Answer the user's questions as requested.

    \item \texttt{Finished(content='')}

    Mark the task as completed and inform the user of the task execution status.
\end{itemize}

\medskip

\#\#\# Instructions

\begin{itemize}
    \item Make sure you understand the task goal to avoid wrong actions.

    \item Make sure you carefully examine the current screenshot. Sometimes the summarized history might not be reliable, over-claiming some effects.

    \item If additional information is needed during task execution, use \texttt{CallUser} to interact with the user.

    \item Consider exploring the screen by using the \texttt{Swipe} action with different directions to reveal additional content.

    \item To copy text: first select the exact text you want to copy, which usually also brings up the text selection bar, then click the \texttt{copy} button in bar.

    \item To paste text into a text box, first long press the text box, then usually the text selection bar will appear with a \texttt{paste} button in it.
\end{itemize}

\medskip

\#\#\# Output Format

\texttt{<think> your thinking process </think>}

\texttt{<action> the next action </action>}

\medskip

\#\#\# User Task

\texttt{\color{red}\{user\_task\}}

\end{tcolorbox}

\subsection{Web}

\begin{tcolorbox}[
    title=Web Prompt,
    breakable
]\label{web_prompt}

\textbf{You are a GUI Browser Agent.}

Your task is to analyze a given user task, review current screenshot and previous actions, and determine the next action to complete the task.

\medskip

\#\#\# Available Actions

You may execute one of the following functions:

\begin{itemize}
    \item \texttt{Click(box=(x1,y1))}

    Perform a tap action at the specified screen coordinate. Valid coordinates range from the top-left corner $(0,0)$ to the bottom-right corner $(999,999)$.

    \item \texttt{Drag(start=(x1,y1), end=(x2,y2))}

    Perform a drag action by long-pressing at the start coordinate for a few seconds and then dragging to the end coordinate. This is typically used for adjusting element layouts, moving sliders, solving slider captchas, etc. Valid coordinates range from $(0,0)$ to $(999,999)$.

    \item \texttt{Scroll(box=(x1, y1), direction='up/down/left/right')}

    Perform a scroll action on coordinate $(x1,y1)$. This is typically used for scrolling to find content, switching tabs, pulling down the notification shade, etc. Valid coordinates range from $(0,0)$ to $(999,999)$, scroll direction=\texttt{'up/down/left/right'}.

    \item \texttt{Type(content='')}

    Enter the specified text into the currently active input field.

    \item \texttt{Launch(url='')}

    Launch the target url. Use this action when the target website is not currently visible on the screen.

    \item \texttt{Wait()}

    Wait for the current page, animation, or content to finish loading.

    \item \texttt{GetUrl()}

    Get the URL of the current browser tab. The URL is returned at the beginning of the next user message.

    \item \texttt{Finished(content='')}

    Mark the task as completed and inform the user of the task execution status.

    \item \texttt{TakeNote(content='')}

    Record important information from screenshots avoiding forgetting.

    \item \texttt{CallUser(content='')}

    Request user takeover or additional information when needed, for example, when there are multiple on-screen options that satisfy the requirement.

    \item \texttt{LongPress(box=(x1,y1), duration=20)}

    Press and hold the specified screen coordinate for \texttt{duration} seconds. \texttt{duration} must be a positive number and defaults to 20 when omitted. This can trigger additional options, such as copy, forward, or delete. Valid coordinates range from $(0,0)$ to $(999,999)$.

    \item \texttt{PressBack()}

    Return to the previous page.

    \item \texttt{PressHome()}

    Return to the browser home page.

    \item \texttt{PressEnter()}

    Perform an Enter key action.

    \item \texttt{Hover(box=(x1,y1))}

    Perform a hover action at the specified screen coordinate. This can be used to reveal additional information or options, such as tooltips or dropdown menus. Valid coordinates range from $(0,0)$ to $(999,999)$.

    \item \texttt{DoubleClick(box=(x1,y1))}

    Perform a double tap action at the specified screen coordinate. Valid coordinates range from $(0,0)$ to $(999,999)$.

    \item \texttt{Hotkey(keys=('ctrl', 'c'))}

    Press combination keys. Keys with comma and wrap each key in single quotes. Do not use more than 3 keys in one \texttt{Hotkey} action. Use \texttt{Hotkey(keys=('ctrl', 'tab'))} for the next browser tab or add \texttt{'shift'} for the previous tab.

    \item \texttt{SelectOption(index=3)}

    Choose an option from the last clicked native HTML select element. Use this only when the current user message provides an explicit native select option list. Do not use keyboard navigation hotkeys for native select dropdowns. For custom dropdowns or visible menu items in the screenshot, click the visible option directly.
\end{itemize}

\medskip

\#\#\# Instruction

\begin{itemize}
    \item Today is \texttt{\{current\_date\}}.

    \item Make sure you understand the task goal to avoid wrong actions.

    \item Make sure you carefully examine the the current screenshot. Sometimes the summarized history might not be reliable, over-claiming some effects.

    \item If additional information is needed during task execution, use \texttt{CallUser} to interact with the user.

    \item Consider exploring the screen by using the \texttt{Scroll} action with different directions to reveal additional content.

    \item Try to use simple language when searching.

    \item If you meet \texttt{ERR\_CONNECTION\_CLOSED} or \texttt{404 NOT FOUND} error, please type the website key word in \texttt{https://www.google.com} to find the correct url.

    \item The official website of cryptpad is \texttt{https://cryptpad.fr/}.

    \item Distinguish textbox from button: never \texttt{Type} into a button. If no textbox is visible, try clicking the search icon first---the input field may appear afterward.

    \item Strictly avoid repeating the same action when the webpage remains unchanged---you may have executed the wrong action. Continuous use of \texttt{Wait()} is also \textbf{not allowed}.
\end{itemize}

\medskip

\# Very Important

\textbf{Take Notes:}
\begin{itemize}
    \item You are forgetful and will forget all information from the current screenshot before you scroll to next one. When you see important information(e.g. partial step info) for completing the task in the current screenshot, RECORD it using \texttt{TakeNote(content='...')} before you scrolling it down.

    \item The information needed for a task is often distributed across multiple pages. Even partial information should be taken note of---do not wait until all information is seen.

    \item Before you take \texttt{scroll} action, make sure you have taken notes for all important information in the current screenshot.
\end{itemize}

\textbf{Apply Filters:}
\begin{itemize}
    \item If filters are available on the page, prioritize using filters for precise searching rather than using the search function for fuzzy searching.
\end{itemize}

\medskip

\#\#\# Output Format

\texttt{<think> your thinking process </think>}

\texttt{<action> the next action </action>}

\medskip

\#\#\# User Task

\texttt{\color{red}\{task\}}

\end{tcolorbox}

\subsection{OS}

\begin{tcolorbox}[
    title=OS Prompt,
    breakable
]\label{os_prompt}
\textbf{You are a GUI Agent.}

Your role is to analyze the user's task, provide clear and accurate answers to their questions, and execute the task with precise actions on a desktop operating system. The password of the computer is \texttt{\color{red}\{sudo\_password\}}.

\medskip

\#\#\# Available Actions

You may execute one of the following functions. Coordinates range from the top-left corner (0, 0) to the bottom-right corner (999, 999).

\begin{itemize}
    \item \texttt{Click(box=(x1, y1))}, or \texttt{Click()}

    Perform a left-click at \texttt{box}, or at the current cursor position when \texttt{box} is omitted.

    \item \texttt{DoubleClick(box=(x1, y1))}, or \texttt{DoubleClick()}

    Perform a double-click (selects a word in text). Use \texttt{box} to move first, or omit it to act at the current cursor position.

    \item \texttt{TripleClick(box=(x1, y1))}, or \texttt{TripleClick()}

    Perform a triple-click (selects a line or the content of a single-line input). Use \texttt{box} to move first, or omit it to act at the current cursor position.

    \item \texttt{RightClick(box=(x1, y1))}, or \texttt{RightClick()}

    Perform a right-click to open a context menu. Use \texttt{box} to move first, or omit it to act at the current cursor position.

    \item \texttt{MiddleClick(box=(x1, y1))}, or \texttt{MiddleClick()}

    Perform a middle-click (for example, open a link in a new tab). Use \texttt{box} to move first, or omit it to act at the current cursor position.

    \item \texttt{Hover(box=(x1, y1))}

    Move the cursor immediately to the coordinate WITHOUT clicking.

    \item \texttt{Drag(end=(x2, y2), start=(x1, y1))}

    Drag to \texttt{end} using a fixed 0.5-second drag. \texttt{start} is optional; omit it to begin from the current cursor position.

    \item \texttt{Swipe(amount=-5, axis='vertical')}

    Scroll at the current cursor position. \texttt{amount} is an integer from -4096 to 4096 and controls magnitude and direction: vertical positive scrolls up and negative scrolls down; horizontal positive scrolls right and negative scrolls left.

    \item \texttt{Type(content='')}

    Type the provided text into the focused field. Each \texttt{\textbackslash n} presses Enter.

    \item \texttt{Hotkey(keys=['ctrl', 'c'], repeat=1)}

    Press 1 to 128 listed keys as a keyboard shortcut. Use \texttt{repeat=N} from 1 to 128 to press the shortcut N times.

    \item \texttt{KeyDown(keys=['shift'])}

    Press 1 to 128 listed keys in order and keep them held across later actions and model turns until a matching \texttt{KeyUp}.

    \item \texttt{KeyUp(keys=['shift'])}

    Release 1 to 128 listed keys in order.

    \item \texttt{MouseDown(box=(x1, y1))}, or \texttt{MouseDown()}

    Optionally move to \texttt{box}, then press and hold the left mouse button across later actions and model turns.

    \item \texttt{MouseUp(box=(x1, y1))}, or \texttt{MouseUp()}

    Optionally move to \texttt{box}, then release the left mouse button.

    \item \texttt{Sequence(actions=[Click(box=(x1, y1)), Hotkey(keys=['ctrl', 's'])])}

    Execute 2 to 32 actions in order as one open-loop model turn. Nested \texttt{Sequence} is not allowed, and \texttt{CallUser} or \texttt{Finished} may appear only as the final action.

    \item \texttt{Wait()}

    Wait for the current page, animation, or content to finish loading.

    \item \texttt{CallUser(content='')}

    Request user takeover or report failure when the task cannot be completed or additional information is required.

    \item \texttt{Finished(content='')}

    Mark the task as completed successfully and optionally report details in \texttt{content}.
\end{itemize}

\medskip

\#\#\# Instructions

\begin{itemize}
    \item Make sure you understand the task goal to avoid wrong actions.

    \item Prefer one atomic action per turn. Use \texttt{Sequence} only when every child action is already known and no intermediate screenshot is needed; its children execute open-loop.

    \item \texttt{KeyDown}, \texttt{KeyUp}, \texttt{MouseDown}, and \texttt{MouseUp} preserve input state across turns. Release held input explicitly when it is no longer needed.

    \item Any \texttt{keys} list may contain at most 128 non-empty key names; each key name is limited to 1,024 characters.

    \item \texttt{Swipe} is the only scrolling action and always scrolls at the current cursor position.

    \item Make sure you carefully examine the current screenshot. Sometimes the summarized history might not be reliable, over-claiming some effects.

    \item To submit/search after typing into a field, end the text with a newline --- \texttt{Type(content='query\textbackslash n')} --- which types the text and presses Enter in one action.

    \item To replace the existing content of an input field, use \texttt{TripleClick} to select it, then \texttt{Type} the new content.

    \item To open a submenu/dropdown, use \texttt{Hover} over the parent item to reveal it, then \texttt{Click} the desired entry.

    \item To use a context menu, \texttt{RightClick} the target to open it, then \texttt{Click} the desired entry.

    \item To hold a modifier during another action, use \texttt{KeyDown}, the target action, and \texttt{KeyUp}. Put them in one \texttt{Sequence} only when no intermediate screenshot is needed.

    \item After launching an app, running a command, downloading, or any slow operation, use \texttt{Wait()} to let it finish before continuing.

    \item To press a key or shortcut several times, use \texttt{repeat}, e.g. \texttt{Hotkey(keys=['down'], repeat=5)} or \texttt{Hotkey(keys=['ctrl', 'z'], repeat=3)}, instead of repeating the action.

    \item Consider exploring the screen by using the \texttt{Swipe} action to scroll and reveal additional content.

    \item Use \texttt{Hotkey} for keyboard shortcuts: copy (\texttt{ctrl+c}), paste (\texttt{ctrl+v}), save (\texttt{ctrl+s}), undo (\texttt{ctrl+z}), find (\texttt{ctrl+f}), etc.

    \item If the task cannot be completed or additional information is needed, use \texttt{CallUser}. Use \texttt{Finished} only after successful completion.
\end{itemize}

\medskip

\#\#\# Output Format

\texttt{<think> your thinking process </think>}

\texttt{<action> the next action </action>}

\medskip

\#\#\# User Task

\texttt{\color{red}\{user\_task\}}

\end{tcolorbox}

\subsection{CAPTCHA}

\begin{tcolorbox}[
    title=CAPTCHA Prompt,
    breakable
]\label{captcha_prompt}

\textbf{You are a GUI Agent.}

Your role is to analyze the user's task, provide clear and accurate answers to
their questions, and execute the task with precise actions.

\medskip

\#\#\# Available Actions

You may execute one of the following functions:

\begin{itemize}
    \item \texttt{Click(box=(x1, y1))}

    Perform a tap action at the specified screen coordinate. Valid coordinates
    range from the top-left corner $(0,0)$ to the bottom-right corner
    $(999,999)$.

    \item \texttt{Drag(start=(x1, y1), end=(x2, y2))}

    Perform a drag action by long-pressing at the start coordinate for a few
    seconds and then dragging to the end coordinate. This is typically used for
    adjusting app layouts, moving sliders, solving slider captchas, etc. Valid
    coordinates range from the top-left corner $(0,0)$ to the bottom-right
    corner $(999,999)$.

    \item \texttt{Swipe(start=(x1, y1), end=(x2, y2))}

    Perform a swipe action by dragging from the start coordinate to the end
    coordinate. This is typically used for scrolling to find content, switching
    tabs, pulling down the notification shade, etc. Valid coordinates range from
    the top-left corner $(0,0)$ to the bottom-right corner $(999,999)$.

    \item \texttt{DoubleClick(box=(x1, y1))}

    Perform a double tap action at the specified screen coordinate. Valid
    coordinates range from the top-left corner $(0,0)$ to the bottom-right
    corner $(999,999)$.

    \item \texttt{LongPress(box=(x1, y1))}

    Perform a long-press action at the specified screen coordinate for a certain
    duration. This can be used to trigger additional options, such as copy,
    forward, delete, etc. Valid coordinates range from the top-left corner
    $(0,0)$ to the bottom-right corner $(999,999)$.

    \item \texttt{Type(content='')}

    Enter the specified text into the currently active input field.

    \item \texttt{LaunchApp(app='')}

    Launch the target app. Use this action when the target app is not currently
    visible on the screen.

    \item \texttt{Wait()}

    Wait for the current page, animation, or content to finish loading.

    \item \texttt{CallUser(content='')}

    Request user takeover or additional information when needed, for example,
    when there are multiple on-screen options that satisfy the requirement.

    \item \texttt{GetScreenshot()}

    Take a screenshot and save it to the device's photo album.

    \item \texttt{PressBack()}

    Return to the previous screen.

    \item \texttt{PressHome()}

    Return to the system home screen.

    \item \texttt{PressEnter()}

    Perform an Enter key action.

    \item \texttt{PressRecent()}

    Open the system recent apps screen.

    \item \texttt{Answer(content='')}

    Answer the user's questions as requested.

    \item \texttt{Finished(content='')}

    Mark the task as completed and inform the user of the task execution status.
\end{itemize}

\medskip

\#\#\# Instructions

\begin{itemize}
    \item Make sure you understand the task goal to avoid wrong actions.

    \item Make sure you carefully examine the current screenshot. Sometimes the
    summarized history might not be reliable, over-claiming some effects.

    \item If additional information is needed during task execution, use
    \texttt{CallUser} to interact with the user.

    \item Consider exploring the screen by using the \texttt{Swipe} action with
    different directions to reveal additional content.

    \item To copy text: first select the exact text you want to copy, which
    usually also brings up the text selection bar, then click the \texttt{copy}
    button in bar.

    \item To paste text into a text box, first long press the text box, then
    usually the text selection bar will appear with a \texttt{paste} button in
    it.
\end{itemize}

\medskip

\#\#\# CAPTCHA-Specific Extension

When the user's task is to pass a CAPTCHA, keep the same GUI-agent action
syntax, but follow these additional CAPTCHA rules:

\begin{itemize}
    \item Treat the screenshot as a visual CAPTCHA challenge. First identify the
    CAPTCHA type and the instruction shown in the image.

    \item Ignore unrelated browser/app controls, close buttons, refresh buttons,
    feedback icons, ads, page chrome, and decorative content unless they are the
    explicit CAPTCHA target.

    \item For CAPTCHA tasks, the effective final action space is
    \texttt{Click}, \texttt{Drag}, \texttt{Type}, and \texttt{LongPress}. Use
    other GUI actions only if the CAPTCHA explicitly requires them.

    \item For selection CAPTCHA tasks, use \texttt{Click} on all target(s)
    implied by the instruction: one \texttt{Click} for a single target, and
    multiple \texttt{Click} actions for multiple targets.

    \item For input CAPTCHA tasks, focus the required field(s) when needed and
    type the complete answer(s).

    \item For manipulation CAPTCHA tasks, use \texttt{Drag} to move, align,
    rotate, or match the visual element to the required final state.

    \item For press-and-hold CAPTCHA tasks, use \texttt{LongPress} on the
    required hold target.

    \item When a CAPTCHA requires multiple interactions, combine all required
    actions in the same \texttt{<action>} tag, separated by commas, in execution
    order.

    \item For CAPTCHA tasks, output the complete action sequence at once. Do not
    output only the first or next action.

    \item Do not add a final submit/confirm click unless the CAPTCHA itself
    explicitly requires that click as part of the solution.
\end{itemize}

CAPTCHA action examples:

\begin{itemize}
    \item Single-target Click:

    \texttt{<action>Click(box=(416,889))</action>}

    \item Multi-target Click:

    \texttt{<action>Click(box=(311,382)),Click(box=(688,579)),}
    \texttt{Click(box=(311,776))</action>}

    \item Text input:

    \texttt{<action>Click(box=(500,889)),Type(content='6')</action>}

    \item Multiple text fields:

    \texttt{<action>Click(box=(385,932)),Type(content='10'),}
    \texttt{Click(box=(514,932)),Type(content='7')</action>}

    \item Drag:

    \texttt{<action>Drag(start=(156,682),end=(485,682))</action>}

    \item Hold button:

    \texttt{<action>LongPress(box=(501,800))</action>}
\end{itemize}

\medskip

\#\#\# Output Format

\texttt{<think> your thinking process </think>}

\texttt{<action> the next action, or the complete CAPTCHA action sequence
</action>}

\medskip

\#\#\# User Task

\texttt{\color{red}Help me pass the CAPTCHA.}

\end{tcolorbox}

\section{Grounding Synthesized Example}
\begin{figure}[htbp]
	\centering
\includegraphics[width=0.99\textwidth]{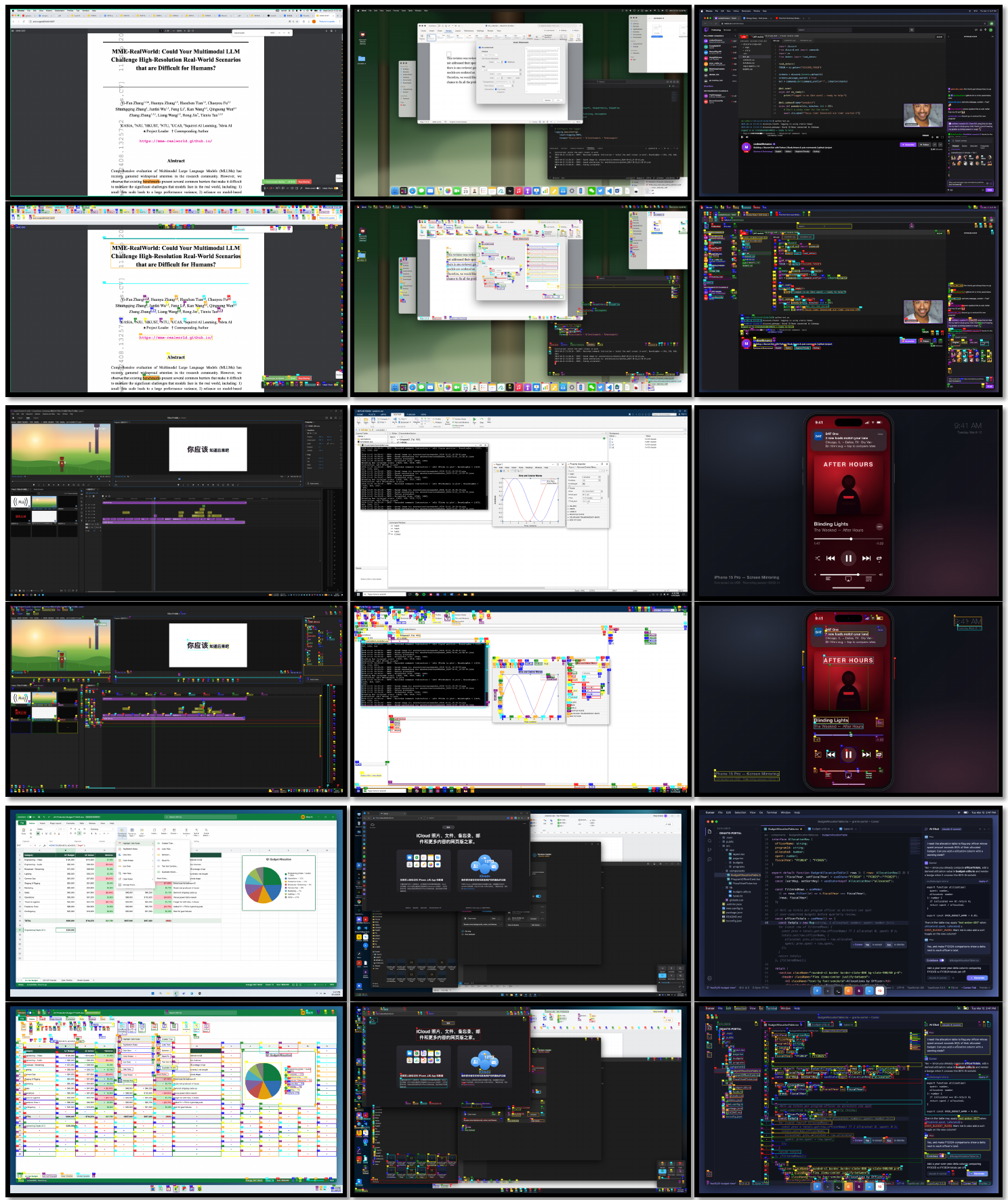}
	\caption{Examples of synthesized GUI Grounding training data. Our pipeline generates diverse and realistic interface screenshots spanning desktop (macOS, Windows), mobile (iOS), and web platforms, covering both professional software and consumer applications. All interfaces are rendered in a real headless Chromium browser via Playwright, ensuring high-fidelity visual output that closely mirrors authentic user environments.}
	\label{fig:guisyn}
\end{figure}

Here, we present some images from the synthetic grounding dataset, featuring various operating systems, office software, entertainment websites, and more, as shown in Figure \ref{fig:guisyn}.

\section{CAPTCHA Benchmarks and Evaluation Protocols}
\label{app:captcha-benchmarks}

\subsection{VenusBench-CAPTCHA}
\label{app:venusbench-captcha}

\paragraph{Motivation.}
CAPTCHAs form a practical bottleneck for autonomous GUI agents: a single
failure at a verification gate can block an otherwise successful workflow.
Solving them is therefore more than an isolated recognition or grounding
problem. A model must interpret the on-screen instruction, extract
fine-grained visual evidence, perform the required semantic, spatial, or
geometric reasoning, and translate its decision into precise executable
actions. Even a correct visual interpretation fails operationally if the
model selects the wrong target, violates the requested order, or executes an
inaccurate click, drag, or text-entry action. VenusBench-CAPTCHA is designed to
stress-test this complete perception--reasoning--action pipeline under a
controlled and reproducible protocol.

\begin{figure*}[t]
  \centering
  \includegraphics[width=\textwidth]{%
    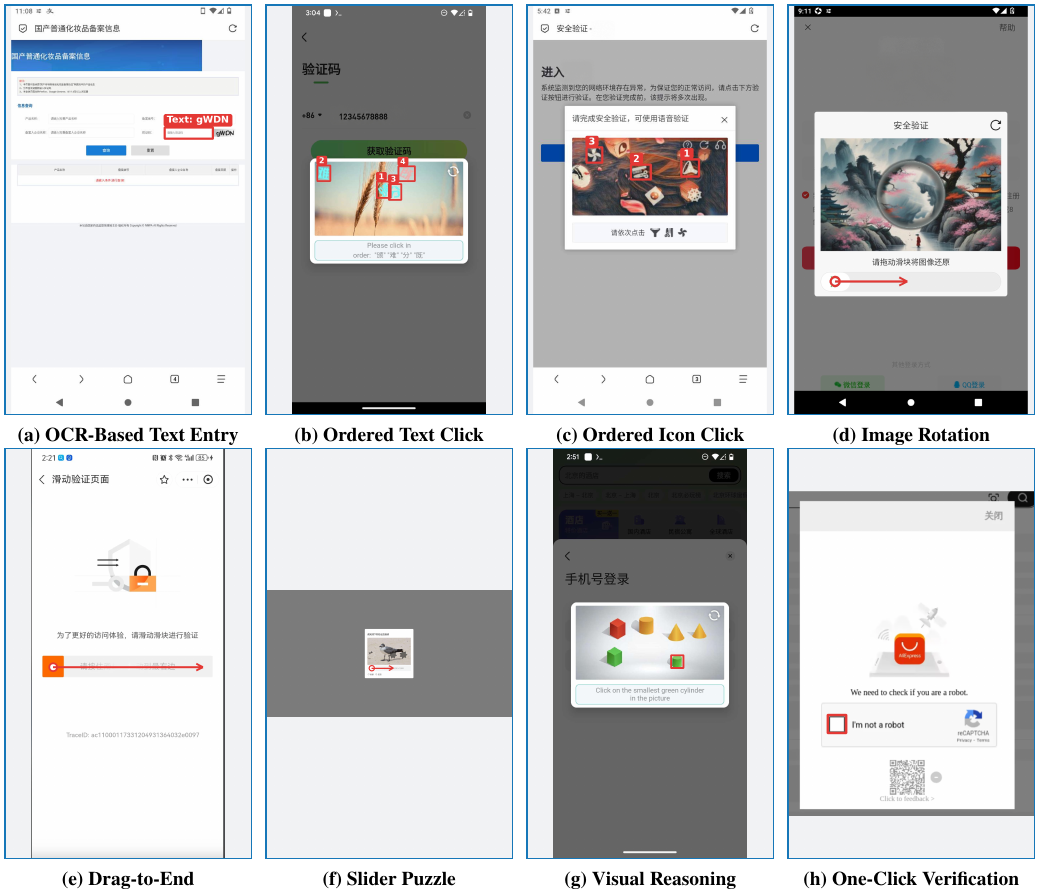}
  \caption{\textbf{Overview of VenusBench-CAPTCHA.}
  Each panel shows one complete, uncropped representative screenshot. The OCR
  label gives the target transcription, numbered boxes indicate the required
  click order, and arrows visualize annotated drag trajectories. These
  annotations are added for presentation only and are not part of the model
  input.}
  \label{fig:venusbench-captcha-overview}
\end{figure*}

\paragraph{Coverage and task design.}
VenusBench-CAPTCHA is a compact evaluation set collected from practical
CAPTCHA deployment scenarios. It contains 219 screenshots spanning eight
interaction types: OCR-based text entry, ordered text click, ordered icon
click, image rotation, drag-to-end, slider puzzle, visual reasoning, and
one-click verification. The first seven categories contain 30 examples each,
while one-click verification contains 9. The captured interfaces retain
surrounding mobile or web context and span 70 image-size configurations across
portrait and landscape layouts, rather than following a single visual
template. As illustrated in
Figure~\ref{fig:venusbench-captcha-overview}, the task taxonomy ranges from
recognition and transcription to semantic target selection, order-sensitive
grounding, visual reasoning, and continuous geometric manipulation. Every task
is mapped to a common executable action language---\texttt{Click}, a
\texttt{Click}--\texttt{Type} pair, or \texttt{Drag}---under the same
CAPTCHA-specific prompt, allowing heterogeneous mechanisms to be compared
through a consistent agent interface.

\paragraph{Diagnostic value.}
The benchmark supports both overall model comparison and category-level
capability diagnosis. Its categories separately probe text recognition and
entry, fine-grained target grounding, sequence following, spatial
transformation, visual reasoning, and drag-distance estimation. Evaluation is
performed on executable actions rather than answer text alone: the predicted
action type, count, target region, order, entered text, and drag geometry are
checked as applicable. This exposes failures that answer-only or single-point
grounding metrics can hide. The fixed screenshots and deterministic evaluator
enable reproducible comparison between general-purpose multimodal models and
GUI-specialized agents, while the compact size makes the benchmark practical
for routine regression testing. The broad performance spread in
Table~\ref{tab:venusbench-captcha-results} further indicates that the benchmark
remains discriminative across the evaluated model families.

\subsection{Public-Benchmark Task Coverage}
\label{app:captcha-public-benchmark-coverage}

In addition to VenusBench-CAPTCHA, we evaluate four public CAPTCHA benchmarks.
These are Spatial-CAPTCHA-Bench~\cite{kharlamova2026spatialcaptcha} and
MCA-Bench~\cite{wu2026mcabench}, together with
NextGen-CAPTCHAs~\cite{liu2026nextgencaptchas} and Open
CaptchaWorld~\cite{luo2025opencaptchaworld}. We describe below their task
coverage and any benchmark-specific selection decisions used in our
evaluation.

\paragraph{Spatial-CAPTCHA-Bench.}
We use the complete benchmark, comprising 1{,}050 image--text instances and
seven task formulations at easy, medium, and hard difficulty levels. These
tasks cover four spatial abilities: reference-system reasoning, orientation
and perspective taking, mental rotation, and multi-step spatial visualization.
No task type or example is excluded.

\paragraph{MCA-Bench.}
MCA-Bench contains a 4{,}000-example test set with 200 examples in each of 20
task categories. The categories span static visual recognition,
point-and-click localization, interactive manipulation, and textual logic
question answering. Because the original paper does not specify a
finer-grained test-set partition, and evaluating all 200 examples per category
for every model would incur substantial inference cost, we randomly select 50
of the 200 examples in each category instead of evaluating the full set. This
procedure yields a fixed, category-balanced subset of 1{,}000 examples that is
shared by all evaluated models.

\paragraph{NextGen-CAPTCHAs.}
We evaluate the following 15 task types:
\begin{quote}
\small\raggedright
\texttt{Dice\_Roll\_Path}, \texttt{Color\_Counting},
\texttt{Hole\_Counting}, \texttt{Rotation\_Match},
\texttt{Backmost\_Layer}, \texttt{Layered\_Stack},
\texttt{Illusory\_Ribbons}, \texttt{Multi\_Script},
\texttt{Box\_Folding}, \texttt{3D\_Viewpoint},
\texttt{Shadow\_Direction}, \texttt{Subway\_Paths},
\texttt{Occluded\_Pattern\_Counting}, \texttt{Shadow\_Plausible}, and
\texttt{Mirror}.
\end{quote}
We exclude \texttt{Red\_Dot} and
\texttt{Static\_Jigsaw} because their released annotations contain label
errors that prevent reliable automatic scoring. The remaining ten task types
are dynamic CAPTCHAs and fall outside the scope of the present evaluation.
Consequently, the reported NextGen-CAPTCHAs results cover only the 15 task
types listed above.

\paragraph{Open CaptchaWorld.}
The fixed evaluation set used in our experiments covers the following 16 task
types:
\begin{quote}
\small\raggedright
\texttt{Bingo}, \texttt{Connect\_icon}, \texttt{Coordinates},
\texttt{Dart\_Count}, \texttt{Dice\_Count}, \texttt{Geometry\_Click},
\texttt{Hold\_Button}, \texttt{Image\_Matching},
\texttt{Image\_Recognition}, \texttt{Object\_Match},
\texttt{Patch\_Select}, \texttt{Path\_Finder}, \texttt{Select\_Animal},
\texttt{Slide\_Puzzle}, \texttt{Unusual\_Detection} and \texttt{Rotation\_Match}.
\end{quote}
We do not evaluate \texttt{Click\_Order}, \texttt{Misleading\_Click},
\texttt{Pick\_Area}, or \texttt{Place\_Dot} because their released annotations
contain label errors.

\subsection{Evaluation Protocol}
\label{app:captcha-evaluation-protocol}

\paragraph{Evaluation sets.}
VenusBench-CAPTCHA is evaluated on all 219 examples,
Spatial-CAPTCHA-Bench on all 1{,}050 examples, MCA-Bench on the 1{,}000-example
category-balanced subset described above, NextGen-CAPTCHAs on 319 examples
from the 15 retained task types, and Open CaptchaWorld on its 16 retained task
types. All evaluation sets are determined before running any
model and then held fixed, so every model is compared on exactly the same
data. Invalid annotations are excluded rather than corrected after observing
model outputs, and no task or example is selected according to model
performance.

\paragraph{Model interface and action parsing.}
We use a single-turn setting: the model receives one CAPTCHA screenshot with
the task instruction and a CAPTCHA-specific prompt, and is instructed to emit
the complete solution in one response. The supported action language consists
of \texttt{Click}, \texttt{LongPress}, \texttt{Click}--\texttt{Type} pairs,
and \texttt{Drag}; only the action sequence is scored, while reasoning text is
ignored. Models may produce normalized coordinates in $[0,999]$ or $[0,1]$,
or absolute pixel coordinates. We map all predicted coordinates back to the
native image resolution before scoring.

\paragraph{Action-level correctness.}
Scoring is strict: the predicted action types and counts must exactly match the
reference, and any missing, extra, malformed, or unsupported action makes the
response incorrect. A \texttt{Click} or \texttt{LongPress} is correct only if
it has the required action type and falls inside a distinct annotated bounding
box. Tasks marked as ordered additionally require the annotated action order;
otherwise, predictions are matched one-to-one to the target boxes without an
order constraint. Text-entry tasks require both a click inside the
corresponding input box and an exact answer match after trimming surrounding
whitespace. A drag task requires exactly one \texttt{Drag}: its start point
must lie inside the annotated handle box, its signed horizontal displacement
must have the correct direction and satisfy the configured relative-error
tolerance, and its end-point vertical error must be below five pixels. For the
multi-answer \texttt{bingo} task, exactly two clicks must match any annotated
valid pair, irrespective of their order.

\end{appendix}

\end{document}